\documentclass[10pt,twocolumn,letterpaper]{article}

\usepackage{cvpr}
\usepackage{times}
\usepackage{epsfig}
\usepackage{graphicx}
\usepackage{amsmath}
\usepackage{amssymb}
\usepackage{float}

\usepackage{bm}
\usepackage{subcaption}
\usepackage{epstopdf}
\usepackage{enumitem}
\usepackage{pifont}
\newcommand{\cmark}{\ding{51}}
\newcommand{\xmark}{\ding{55}}
\newcommand{\FigTabBelowcaptionskip}{\setlength{\belowcaptionskip}{-0.5cm}}
\newcommand{\FigTabBeforecaptionskip}{\vspace{-5pt}}

\usepackage[breaklinks=true,bookmarks=false]{hyperref}

\cvprfinalcopy 

\def\cvprPaperID{387} 

\begin{document}

\title{3D Hand Shape and Pose Estimation from a Single RGB Image}

\author{Liuhao Ge$^{1}$\thanks{This work was done when Liuhao Ge was a research intern at Snap Inc.},~ Zhou Ren$^{2}$, Yuncheng Li$^{3}$, Zehao Xue$^{3}$, Yingying Wang$^{3}$, Jianfei Cai$^{1}$, Junsong Yuan$^{4}$\\
	\!\!\!\!\!\!\!\!\!\!\!\!\!\!\!\!\!\!\!\!\!\!\!$^1$Nanyang Technological University~~~~~~~~~~~~~~~$^2$Wormpex AI Research\\
	~~~~~~~~~~~~~~~~~$^3$Snap Inc.~~~~~~~~~~~~~~~~~~~~$^4$State University of New York at Buffalo\\
	{\tt\small ge0001ao@e.ntu.edu.sg, zhou.ren@bianlifeng.com, yuncheng.li@snap.com,}\vspace{-3pt}\\
	{\tt\small zehao.xue@snap.com, ywang@snap.com, asjfcai@ntu.edu.sg, jsyuan@buffalo.edu}
}


\maketitle

\begin{abstract}
	This work addresses a novel and challenging problem of estimating the full 3D hand shape and pose from a single RGB image. Most current methods in 3D hand analysis from monocular RGB images only focus on estimating the 3D locations of hand keypoints, which cannot fully express the 3D shape of hand. In contrast, we propose a Graph Convolutional Neural Network (Graph CNN) based method to reconstruct a full 3D mesh of hand surface that contains richer information of both 3D hand shape and pose. To train networks with full supervision, we create a large-scale synthetic dataset containing both ground truth 3D meshes and 3D poses. When fine-tuning the networks on real-world datasets without 3D ground truth, we propose a weakly-supervised approach by leveraging the depth map as a weak supervision in training. Through extensive evaluations on our proposed new datasets and two public datasets, we show that our proposed method can produce accurate and reasonable 3D hand mesh, and can achieve superior 3D hand pose estimation accuracy when compared with state-of-the-art methods.

\end{abstract}

\section{Introduction}
\label{Introduction}

Vision-based 3D hand analysis is a very important topic because it has many applications in virtual reality (VR) and augmented reality (AR). However, despite years of \mbox{studies}~\cite{rehg1994visual,wu2001hand,wu2005analyzing,stenger2006model,sridhar2016real,ge2017_3D,Liang2019Hough}, it remains an open problem due to the diversity and complexity of hand shape, pose, gesture, occlusion, \textit{etc}. In the past decade, we have witnessed a rapid advance in 3D hand pose estimation from depth images~\cite{oikonomidis2011efficient,tompson2014real,ge2016Robust,Ge2018Robust,Ge2018Real,yuan2018depth,ge2018hand,ge2018point}. Considering \mbox{RGB} cameras are more widely available than depth cameras, some \mbox{recent} works start looking into 3D hand analysis from monocular \mbox{RGB} images, and mainly focus on estimating sparse 3D hand joint locations but ignore dense 3D hand shape~\cite{zimmermann2017learning,spurr2018cross,mueller2018ganerated,cai2018weakly,iqbal2018hand,panteleris2018using,rad2018domain}. However, many immersive VR and AR applications often require accurate estimation of both 3D hand pose and 3D hand shape.

\begin{figure}[t]
	\FigTabBelowcaptionskip
	\begin{center}
		\includegraphics[width=0.96\linewidth]{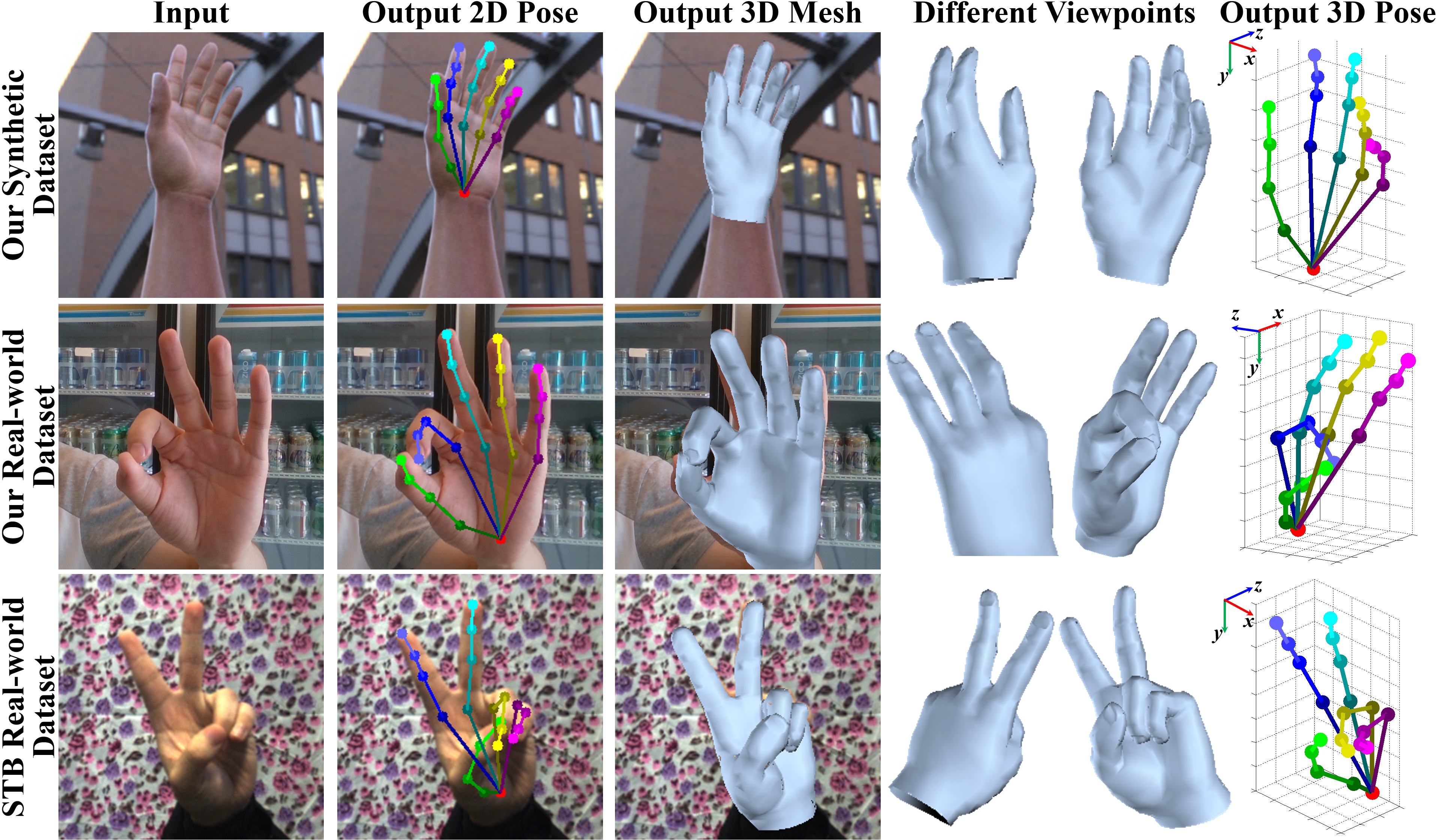}
		\FigTabBeforecaptionskip	
		\caption{Our proposed method is able to not only estimate 2D/3D hand joint locations, but also recover a full 3D mesh of hand surface from a single RGB image. We show our estimation results on our proposed synthetic and real-world datasets as well as the STB real-world dataset~\cite{zhang20163d}.}
		\label{fig:Fig_Visualization}
	\end{center}
\end{figure}

This motivates us to bring out a more challenging task: \emph{how to jointly estimate not only the 3D hand joint locations, but also the full 3D mesh of hand surface from a single RGB image?} In this work, we develop a sound solution to this task, as illustrated in Fig.~\ref{fig:Fig_Visualization}.

The task of single-view 3D hand shape estimation has been studied previously, but mostly in controlled settings, where a depth sensor is available. The basic idea is to fit a generative 3D hand model to the input depth image with iterative optimization~\cite{taylor2014user,makris2015model,Khamis2015learning,joseph2016fits,tkach2017online,remelli2017low}. In contrast, here we consider to estimate 3D hand shape from a monocular \mbox{RGB} image, which has not been extensively studied yet. The absence of explicit depth cues in RGB images makes this task difficult to be solved by iterative optimization approaches. In this work, we apply deep neural networks that are trained in an end-to-end manner to recover 3D hand mesh directly from a single RGB image. Specifically, we predefine the topology of a triangle mesh representing the hand surface, and aim at estimating the 3D coordinates of all the vertices in the mesh using deep neural networks. To achieve this goal, there are several challenges.

The first challenge is the high dimensionality of the output space for 3D hand mesh generation. Compared with estimating sparse 3D joint locations of the hand skeleton (\textit{e.g.}, 21 joints), it is much more difficult to estimate 3D coordinates of dense mesh vertices (\textit{e.g.}, 1280 vertices) using conventional CNNs. One straightforward solution is to follow the common approach used in human body shape estimation~\cite{tung2017self,tan2017indirect,pavlakos2018learning,kanazawa2018end}, namely to regress low-dimensional parameters of a predefined deformable hand model, \textit{e.g.}, MANO~\cite{romero2017embodied}.

In this paper we argue that the output 3D hand mesh vertices in essence are graph-structured data, since a 3D mesh can be easily represented as a graph. To output such graph-structured data and better exploit the topological relationship among mesh vertices in the graph, motivated by \mbox{recent} works on Graph \mbox{CNNs}~\cite{defferrard2016convolutional,ranjan2018generating,wang2018pixel2mesh}, we propose a novel Graph CNN-based approach. Specifically, we adopt graph convolutions \cite{defferrard2016convolutional} hierarchically with upsampling and nonlinear activations to generate 3D hand mesh vertices in a graph from image features which are extracted by backbone \mbox{networks}. With such an end-to-end trainable framework, our Graph CNN-based method can better represent the highly variable 3D hand shapes, and can better express the local details of 3D hand shapes.
Besides the computational model, an additional challenge is the lack of ground truth 3D hand mesh training data for real-world images. Manually annotating the ground truth 3D hand meshes on real-world \mbox{RGB} images is extremely laborious and time-consuming. We thus choose to create a large-scale synthetic dataset containing the ground truth of both 3D hand mesh and 3D hand pose for training. However, models trained on the synthetic dataset usually produce unsatisfactory estimation results on real-world datasets due to the domain gap between them. To address this issue, inspired by~\cite{cai2018weakly,pavlakos2018learning}, we propose a novel weakly-supervised method by leveraging depth map as a weak supervision for 3D mesh generation, since depth map can be easily captured by an \mbox{RGB-D} camera when collecting real-world training data. More specifically, when fine-tuning on real-world datasets, we render the generated 3D hand mesh to a depth map on the image plane and minimize the depth map loss against the reference depth map, as shown in Fig.~\ref{fig:Fig_Framework}. Note that, during testing, we only need an RGB image as input to estimate full 3D hand shape and pose.
To the best of our knowledge, we are the first to handle the problem of estimating not only 3D hand pose but also full 3D hand shape from a single RGB image. Our main contributions are summarized as follows:
\begin{itemize}[leftmargin=*]\itemsep0pt
	\item We propose a novel end-to-end trainable hand mesh generation approach based on Graph CNN~\cite{defferrard2016convolutional}. Experiments show that our method can well represent hand shape variations and capture local details. Furthermore, we observe that by estimating full 3D hand mesh, our method boost the accuracy performance of 3D hand pose estimation, as validated in Sec.~\ref{Experiments:eval_pose}.
	\item We propose a weakly-supervised training pipeline on real-world dataset, by rendering the generated 3D mesh to a depth map on the image plane and leveraging the reference depth map as a weak supervision, without requiring any annotations of 3D hand mesh or 3D hand pose for real-world images.
	\item We introduce the first large-scale synthetic RGB-based 3D hand shape and pose dataset as well as a small-scale real-world dataset, which contain the annotation of both 3D hand joint locations and the full 3D meshes of hand surface. We will share our datasets publicly upon the acceptance of this work.
\end{itemize}

We conduct comprehensive experiments on our proposed synthetic and real-world datasets as well as two public datasets~\cite{zhang20163d,zimmermann2017learning}. Experimental results show that our proposed method can produce accurate and reasonable 3D hand mesh with real-time speed on GPU, and can achieve superior accuracy performance on 3D hand pose estimation when compared with state-of-the-art methods.

\section{Related Work}
\label{Related_Work}
\noindent\textbf{3D hand shape and pose estimation from depth images}: Most previous methods estimate 3D hand shape and pose from depth images by fitting a deformable hand model to the input depth map with iterative optimization~\cite{taylor2014user,makris2015model,Khamis2015learning,joseph2016fits,tkach2017online,remelli2017low}. A recent method~\cite{malik2018deephps} was proposed to estimate pose and shape parameters from the depth image using \mbox{CNNs}, and recover 3D hand meshes using \mbox{LBS}. The CNNs are trained in an end-to-end manner with mesh and pose losses. However, the quality of their recovered hand meshes is restricted by their simple LBS model.
\vspace{2pt}

\noindent\textbf{3D hand pose estimation from RGB images}: Pioneering works~\cite{wu2005analyzing,de2011model} estimate hand pose from RGB image sequences.  Gorce et al.~\cite{de2011model} proposed estimating 3D hand pose, the hand texture and the illuminant dynamically through minimization of an objective function. 
Sridhar et al.~\cite{sridhar2013interactive} adopted multi-view RGB images and depth data to estimate the 3D hand pose by combining a discriminative method with local optimization. With the advance of deep learning and the wide applications of monocular RGB cameras, many recent works estimate 3D hand pose from a single RGB image using deep neural \mbox{networks} \cite{zimmermann2017learning,spurr2018cross,mueller2018ganerated,cai2018weakly,iqbal2018hand,rad2018domain}. However, few works focus on 3D hand shape estimation from RGB images. Panteleris \textit{et al}. \cite{panteleris2018using} proposed to fit a 3D hand model to the estimated 2D joint locations. But the hand model is controlled by 27 hand pose parameters, thus it cannot well adapt to various hand shapes. In addition, this method is not an end-to-end framework for generating 3D hand mesh.
\vspace{2pt}

\noindent\textbf{3D human body shape and pose estimation from a single RGB image}: Most recent methods rely on SMPL, a body shape and pose model~\cite{loper2015smpl}. Some methods fit the SMPL model to the detected 2D keypoints~\cite{bogo2016keep,lassner2017unite}. Some methods regress SMPL parameters using CNNs with supervisions of silhouette and/or 2D keypoints~\cite{tan2017indirect,pavlakos2018learning,kanazawa2018end}. A more recent method~\cite{varol2018bodynet} predicts a volumetric representation of human body. Different from these methods, we propose to estimate 3D mesh vertices using Graph CNNs in order to learn nonlinear hand shape variations and better utilize the relationship among vertices in the mesh topology. In addition, instead of using 2D silhouette or 2D keypoints to weakly supervise the network training, we propose to leverage the depth map as a weak 3D supervision when training on real-world datasets without 3D mesh or 3D pose annotations.


\section{3D Hand Shape and Pose Dataset Creation}
\label{Methodology:dataset}
Manually annotating the ground truth of 3D hand meshes and 3D hand joint locations for real-world RGB images is extremely laborious and time-consuming. To overcome the \mbox{difficulties} in real-world data annotation, some \mbox{works}~\cite{simon2017hand,zimmermann2017learning,mueller2017real} have adopted synthetically generated hand RGB images for training. However, existing hand \mbox{RGB} image datasets~\cite{simon2017hand,zhang20163d,zimmermann2017learning,mueller2017real} only provide the annotations of 2D/3D hand joint locations, and they do not contain any 3D hand shape annotations. Thus, these datasets are not suitable for the training of the 3D hand shape estimation task.

\begin{figure}[t]
	\setlength{\belowcaptionskip}{-0.6cm}
	\begin{center}
		\includegraphics[width=0.938\linewidth]{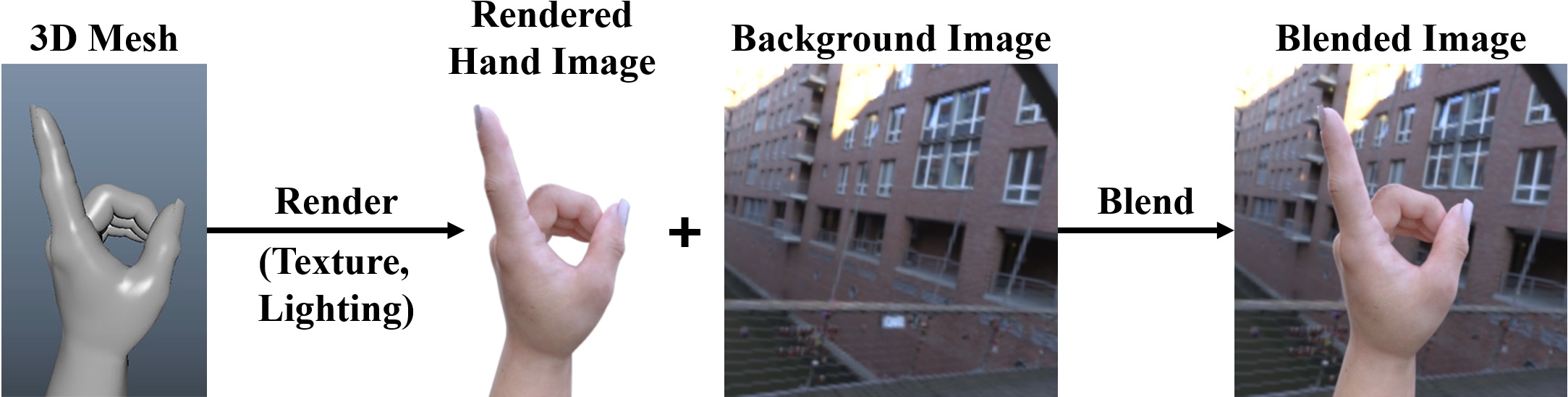}
		\FigTabBeforecaptionskip	
		\caption{Illustration of our synthetic hand shape and pose dataset creation as well as background image augmentation during training.}
		\label{fig:Fig_hand_render_dataset}
	\end{center}
\end{figure}

\begin{figure*}[t]
	\FigTabBelowcaptionskip
	\begin{center}
		\includegraphics[width=0.85\linewidth]{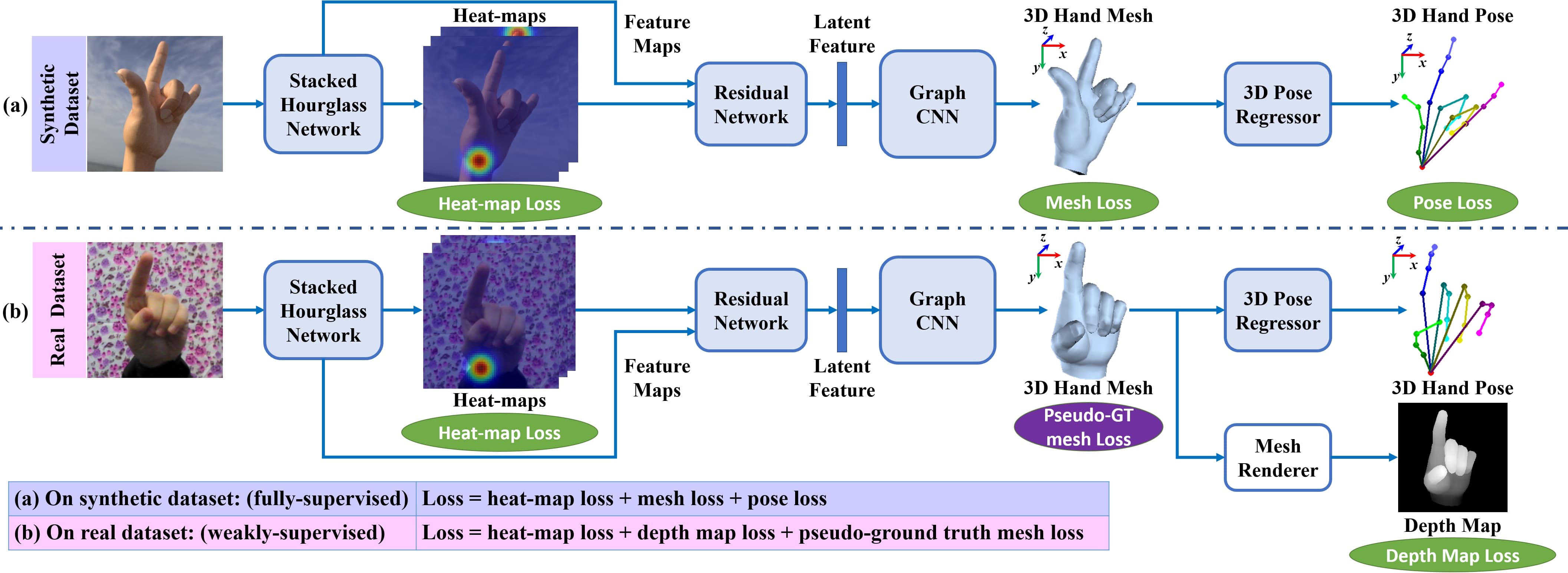}
		\FigTabBeforecaptionskip
		\caption{Overview of our method for 3D hand shape and pose estimation from a single RGB image. Our network model is first trained on a synthetic dataset in a fully supervised manner with heat-map loss, 3D mesh loss, and 3D pose loss, as shown in (a); and then fine-tuned on a real-world dataset without 3D mesh or 3D pose ground truth in a weakly-supervised manner by innovatively introducing a pseudo-ground truth mesh loss and a depth map loss, as shown in (b). For both (a) and (b), the input RGB image is first passed through a two-stacked hourglass network~\cite{newell2016stacked} for extracting feature maps and 2D heat-maps, which are then combined and encoded as a latent feature vector by a residual network~\cite{he2016deep}. The latent feature is fed into a Graph CNN~\cite{defferrard2016convolutional} to infer the 3D coordinates of mesh vertices. Finally, the 3D hand pose is linearly regressed from the 3D hand mesh. During training on the real-world dataset, as shown in (b), the generated 3D hand mesh is rendered to a depth map to compute the depth map loss against the reference depth map. Note that this step is not involved in testing.}
		\label{fig:Fig_Framework}
	\end{center}
\end{figure*}

In this work, we create a large-scale synthetic hand shape and pose dataset that provides the annotations of both 3D hand joint locations and full 3D hand meshes. In particular, we use Maya~\cite{Maya} to create a 3D hand model and rig it with joints, and then apply photorealistic textures on it as well as natural lighting using High-Dynamic-Range (HDR) images. We model hand variations by creating blend shapes with different shapes and ratios, then applying random weights on the blend shapes. To fully explore the pose space, we create hand poses from 500 common hand gestures and 1000 unique camera viewpoints. To simulate real-world diversity, we use 30 lightings and five skin colors. We render the hand using global illumination with off-the-shelf Arnold renderer~\cite{Arnold}. The rendering tasks are distributed onto a cloud render farm for maximum efficiency. In total, our synthetic dataset contains 375,000 hand RGB images with large \mbox{variations}. We use 315,000 images for training and 60,000 images for validation. During training, we randomly sample and crop background images from COCO~\cite{lin2014microsoft}, \mbox{LSUN}~\cite{yu15lsun}, and \mbox{Flickr}~\cite{Flickr} datasets, and blend them with the rendered hand images, as shown in Fig.~\ref{fig:Fig_hand_render_dataset}.

In addition, to quantitatively evaluate the performance of hand mesh estimation on real-world image, we create a real-world dataset containing 583 hand RGB images with the annotations of 3D hand mesh and 3D hand joint locations. To facilitate the 3D annotation, we capture the corresponding depth images using an Intel RealSense RGB-D \mbox{camera}~\cite{RealSense} and manually adjust the 3D hand model in Maya with the reference of both RGB images and depth points. In this work, this real-world dataset is only used for evaluation.

\section{Methodology}
\label{Methodology}

\subsection{Overview}
\label{Methodology:Overview}
We propose to generate a full 3D mesh of the hand surface and the 3D hand joint locations directly from a single monocular RGB image, as illustrated in Fig.~\ref{fig:Fig_Framework}. Specifically, the input is a single RGB image centered on a hand, which is passed through a two-stacked hourglass network~\cite{newell2016stacked} to infer 2D heat-maps. The estimated 2D heat-maps, combined with the image feature maps, are encoded as a latent feature vector by using a residual network~\cite{he2016deep} that contains eight residual layers and four max pooling layers. The encoded latent feature vector is then input to a Graph \mbox{CNN}~\cite{defferrard2016convolutional} to infer the 3D coordinates of ${N}$ vertices ${\mathcal{V} = \left\{ {{{\bm{v}}_i}} \right\}_{i = 1}^N}$ in the 3D hand mesh. The 3D hand joint locations ${{\bf{\Phi}}  = \left\{ {{{\bm{\phi}} _j}} \right\}_{j = 1}^J}$ are linearly regressed from the reconstructed 3D hand mesh vertices by using a simplified linear Graph CNN.

In this work, we first train the network models on a synthetic dataset and then fine-tune them on real-world \mbox{datasets}. On the synthetic dataset that contains the ground truth of 3D hand meshes and 3D hand joint locations, we train the networks end-to-end in a fully-supervised manner by using 2D heat-map loss, 3D mesh loss, and 3D pose loss. More details will be presented in Section~\ref{Methodology:Training_synth}. On the real-world dataset, the \mbox{networks} can be fine-tuned in a weakly-supervised manner without requiring the ground truth of 3D hand meshes or 3D hand joint locations. To achieve this target, we leverage the reference depth map available in training, which can be easily captured from a depth camera, as a weak supervision during the fine-tuning, and employ a differentiable renderer to render the generated 3D mesh to a depth map from the camera viewpoint. To guarantee the mesh quality, we generate the pseudo-ground truth mesh from the pretrained model as an additional supervision. More details will be presented in Section~\ref{Methodology:Training_real}.

\subsection{Graph CNNs for Mesh and Pose Estimation}
\label{Methodology:Graph_cnns}
Graph CNNs have been successfully applied in modeling graph structured data~\cite{wang2018pixel2mesh,yan2018spatial,verma2018feastnet}. As 3D hand mesh is of graph structure by nature, in this work we adopt the Chebyshev Spectral Graph CNN~\cite{defferrard2016convolutional} to generate 3D coordinates of vertices in the hand mesh and estimate 3D hand pose from the generated mesh.

\begin{figure}[t]
	\FigTabBelowcaptionskip
	\begin{center}
		\includegraphics[width=1.0\linewidth]{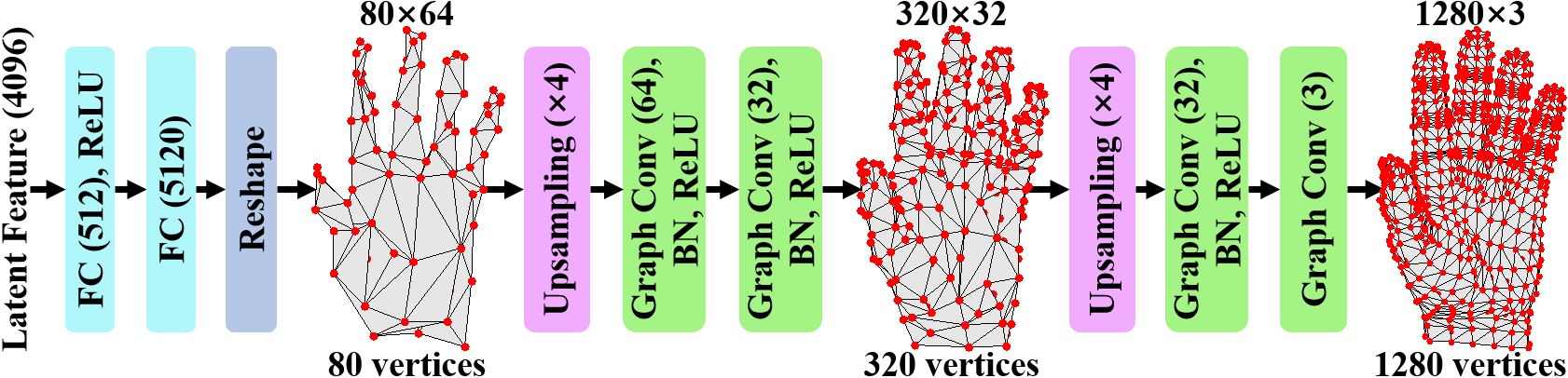}	
		\vspace{-15pt}
		\caption{Architecture of the Graph CNN for mesh generation. The input is a latent feature vector extracted from the input RGB image. Passing through two fully-connected (FC) layers, the feature vector is transformed into 80 vertices with 64-dim features in a coarse graph. The features are upsampled and allocated to a finer graph. With two upsampling layers and four graph convolutional \mbox{layers}, the network outputs 3D coordinates of the 1280 mesh vertices. The numbers in parentheses of FC layers and graph convolutions represent the dimensions of output features.}
		\label{fig:fig_graphcnn}
	\end{center}
\end{figure}
A 3D mesh can be represented by an undirected graph ${\mathcal{M} = \left( {\mathcal{V},\mathcal{E},{W}} \right)}$, where ${\mathcal{V} = \left\{ {{{\bm{v}}_i}} \right\}_{i = 1}^N}$ is a set of ${N}$ vertices in the mesh, ${\mathcal{E} = \left\{ {{{\bm{e}}_i}} \right\}_{i = 1}^E}$ is a set of ${E}$ edges in the mesh, ${{{W}} = {\left( {{w_{ij}}} \right)_{N \times N}}}$ is the adjacency matrix, where ${w_{ij} = 0}$ if ${\left( {i,j} \right) \notin \mathcal{E}}$, and ${w_{ij} = 1}$ if ${\left( {i,j} \right) \in \mathcal{E}}$. The normalized graph Laplacian~\cite{chung1997spectral} is computed as ${L = {I_N} - {D^{ - {1 \mathord{\left/
					{\vphantom {1 2}} \right.
					\kern-\nulldelimiterspace} 2}}}W{D^{ - {1 \mathord{\left/
					{\vphantom {1 2}} \right.
					\kern-\nulldelimiterspace} 2}}}}$, where ${D = {\rm{diag}}\left( {\sum\nolimits_j {{w_{ij}}} } \right)}$ is the diagonal degree matrix, ${I_N}$ is the identity matrix. Here, we assume that the topology of the triangular mesh is fixed and is predefined by the hand mesh model, \textit{i.e.}, the adjacency matrix ${W}$ and the graph Laplacian ${L}$ of the graph ${\mathcal{M}}$ are fixed during training and testing.


Given a signal ${{\bf{f}} = {\left( {{f_1}, \cdots ,{f_N}} \right)^T} \in {{\mathbb{R}}^{N \times F}}}$ on the vertices of graph ${\mathcal{M}}$, it represents ${F}$-dim features of ${N}$ vertices in the 3D mesh. In Chebyshev Spectral Graph \mbox{CNN}~\cite{defferrard2016convolutional}, the graph convolutional operation on a graph signal ${{{\bf{f}}_{\rm{in}}} \in {{\mathbb{R}}^{N \times F_{\rm{in}}}}}$ is defined as
\begin{equation}
{{\bf{f}}_{\rm{out}}} = \sum\nolimits_{k = 0}^{K - 1} {{T_k}\left( {\tilde L} \right) \cdot {{\bf{f}}_{\rm{in}}} \cdot {\theta _k}},
\label{eq:Eq_graph_conv}
\end{equation}
where ${{T_k}\left( x  \right) = 2x{T_{k - 1}}\left( x  \right) - {T_{k - 2}}\left( x  \right)}$ is the Chebyshev polynomial of degree ${k}$, ${{T_0}=1}$, ${{T_1}=x}$; ${\tilde L \in {{\mathbb{R}}^{N \times N}}}$ is the rescaled Laplacian, ${\tilde L = {{2L} \mathord{\left/
		{\vphantom {{2L} {{\lambda _{\max }}}}} \right.
		\kern-\nulldelimiterspace} {{\lambda _{\max }}}} - {I_N}}$, ${\lambda _{\max }}$ is the maximum eigenvalue of ${L}$; ${\theta _k \in {{\mathbb{R}}^{F_{\rm{in}} \times F_{\rm{out}}}}}$ are the trainable parameters in the graph convolutional layer; ${{\bf{f}}_{\rm{out}} \in {{\mathbb{R}}^{N \times F_{\rm{out}}}}}$ is the output graph signal. This operation is ${K}$-localized \mbox{since} Eq.~\ref{eq:Eq_graph_conv} is a ${K}$-order polynomial of the graph Laplacian, and it only affects the ${K}$-hop neighbors of each central node. Readers are referred to~\cite{defferrard2016convolutional} for more details.

\begin{figure}[t]
	\FigTabBelowcaptionskip
	\begin{center}
		\includegraphics[width=1.0\linewidth]{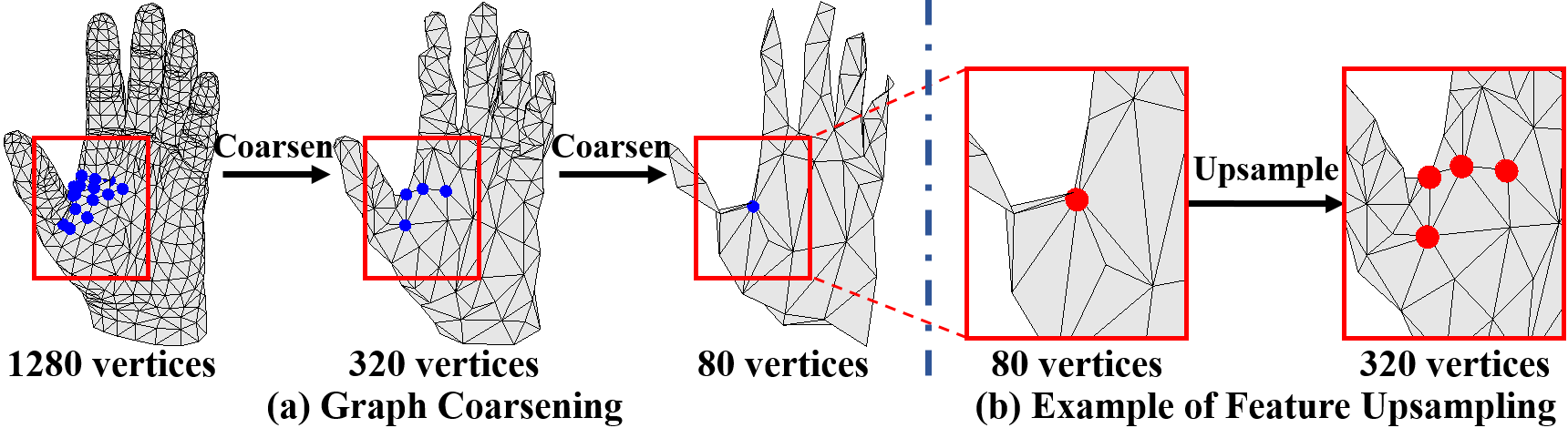}
		\FigTabBeforecaptionskip
		\caption{(a) Given our predefined mesh topology, we first perform graph coarsening~\cite{defferrard2016convolutional} to cluster meaningful {neighborhoods} on graphs and create a tree structure to store correspondences of vertices in graphs at adjacent coarsening {levels}. (b) During the forward propagation, we perform feature upsampling. The feature of a vertex in the coarse graph is allocated to its children vertices in the finer graph.}
		\label{fig:fig_coarsening}
	\end{center}
\end{figure}

In this work, we design a hierarchical architecture for mesh generation by performing graph convolution on graphs from coarse to fine, as shown in Fig.~\ref{fig:fig_graphcnn}. The topologies of coarse graphs are precomputed by graph coarsening, as shown in Fig.~\ref{fig:fig_coarsening} (a), and are fixed during training and testing. Following Defferrard \textit{et al.} \cite{defferrard2016convolutional}, we use the Graclus multilevel clustering {algorithm}~\cite{dhillon2007weighted} to coarsen the graph, and create a tree structure to store correspondences of vertices in graphs at adjacent coarsening levels. During the forward propagation, we upsample features of vertices in the coarse graph to corresponding children vertices in the fine graph, as shown in Fig.~\ref{fig:fig_coarsening} (b). Then, we perform the graph convolution to update features in the graph. All the graph convolutional {filters} have the same support of ${K = 3}$. To make the network output irrelevant to the camera intrinsic \mbox{parameters}, we design the network to output UV coordinates on input image and depth of vertices in the mesh, which can be converted to 3D coordinates in the camera coordinate system using the camera intrinsic matrix. Similar to \cite{zimmermann2017learning,cai2018weakly,spurr2018cross}, we estimate scale-invariant and root-relative depth of mesh vertices.

Considering that 3D joint locations can be estimated directly from the 3D mesh vertices using a linear regressor~\cite{loper2015smpl,romero2017embodied}, we adopt a simplified Graph \mbox{CNN}~\cite{defferrard2016convolutional} with two pooling layers and without nonlinear activation to linearly regress the scale-invariant and root-relative 3D hand joint locations from 3D coordinates of hand mesh vertices.

%

\subsection{Fully-supervised Training on Synthetic Dataset}
\label{Methodology:Training_synth}
We first train the \mbox{networks} on our synthetic hand shape and pose dataset in a fully-supervised manner. As shown in Fig.~\ref{fig:Fig_Framework} (a), the networks are supervised by heat-map loss ${{\mathcal{L}}_{\mathcal{H}}}$, mesh loss ${{\mathcal{L}}_{\mathcal{M}}}$, and 3D pose loss ${{\mathcal{L}}_{\mathcal{J}}}$.

\textbf{Heat-map Loss.} ${{{\mathcal{L}}_{\mathcal{H}}} = \sum\nolimits_{j = 1}^J {\left\| {{\mathcal{H}}_j - {{\hat {\mathcal{H}}}_j}} \right\|_2^2} }$, where ${{\mathcal{H}}_j}$ and ${\hat {\mathcal{H}}_j}$ are the ground truth and estimated heat-maps, respectively. We set the heat-map resolution as 64$\times$64 px. The ground truth heat-map is defined as a 2D Gaussian with a standard deviation of 4 px centered on the ground truth 2D joint location.

\textbf{Mesh Loss.} Similar to~\cite{wang2018pixel2mesh}, ${{\mathcal{L}}_{\mathcal{M}}} = {\lambda _v}{{\mathcal{L}}_{v}} + {\lambda _n}{{\mathcal{L}}_{n}} + {\lambda _e}{{\mathcal{L}}_{e}} + {\lambda _l}{{\mathcal{L}}_{l}}$ is composed of vertex loss ${{\mathcal{L}}_{v}}$, normal loss ${{\mathcal{L}}_{n}}$, edge loss ${{\mathcal{L}}_{e}}$, and Laplacian loss ${{\mathcal{L}}_{l}}$. The {vertex loss} ${{\mathcal{L}}_{v}}$ is to constrain 2D and 3D locations of mesh vertices:
\begin{equation}
{{\mathcal{L}}_{v}} = \sum\nolimits_{i = 1}^N {\left\| {{\bm{v}}_i^{3D} - \hat {\bm{v}}_i^{3D}} \right\|_2^2 + \left\| {{\bm{v}}_i^{2D} - \hat {\bm{v}}_i^{2D}} \right\|_2^2},
\end{equation}
where ${{\bm{v}}_i}$ and ${\hat {\bm{v}}_i}$ denote the ground truth and estimated 2D/3D locations of the mesh vertices, respectively. The {normal loss} ${{\mathcal{L}}_{n}}$ is to enforce surface normal consistency:
\begin{equation}
{{\mathcal{L}}_{n}} = \sum\nolimits_{t} {\sum\nolimits_{{\left( {i,j} \right)} \in {t}} {\left\| {\left\langle {\hat {\bm{v}}_i^{3D} - \hat {\bm{v}}_j^{3D},{{\bm{n}}_{t}}} \right\rangle } \right\|_2^2} },
\end{equation}
where ${t}$ is the index of triangle faces in the mesh; ${\left( {i,j} \right)}$ are the indices of vertices that compose one edge of triangle ${t}$; and ${{\bm{n}}_{t}}$ is the ground truth normal vector of triangle face ${t}$, which is computed from ground truth vertices. The {edge loss} ${{\mathcal{L}}_{e}}$ is introduced to enforce edge length consistency:
\begin{equation}
{{\mathcal{L}}_{e}} = \sum\nolimits_{i = 1}^E {{{\left( {\left\| {{{\bm{e}}_i}} \right\|_2^2 - \left\| {{{\hat {\bm{e}}}_i}} \right\|_2^2} \right)}^2}},
\end{equation}
where ${{\bm{e}}_i}$ and ${\hat {\bm{e}}_i}$ denote the ground truth and estimated edge vectors, respectively. The {Laplacian loss} ${{\mathcal{L}}_{l}}$ is introduced to preserve the local surface smoothness of mesh:
\begin{equation}
{{\mathcal{L}}_{l}} = \sum\nolimits_{i = 1}^N {\left\| {{{\bm{\delta}} _i} - {{\sum\nolimits_{{{\bm{v}}_k} \in {\mathcal{N}}\left( {{{\bm{v}}_i}} \right)} {{{\bm{\delta}} _k}} } \mathord{\left/
				{\vphantom {{\sum\nolimits_{k \in N\left( {{v_i}} \right)} {{{\bm{\delta}} _k}} } {B_i}}} \right.
				\kern-\nulldelimiterspace} {B_i}}} \right\|_2^2},
\end{equation}
where ${{{\bm{\delta}} _i} = {\bm{v}}_i^{3D} - \hat {\bm{v}}_i^{3D}}$ is the offset from the estimation to the ground truth, ${{\mathcal{N}}\left( {{{\bm{v}}_i}} \right)}$ is the set of neighboring vertices of ${{{\bm{v}}_i}}$, and ${B_i}$ is the number of vertices in the set ${{\mathcal{N}}\left( {{{\bm{v}}_i}} \right)}$. This loss function prevents the neighboring vertices from having opposite offsets, thus making the estimated 3D hand surface mesh smoother. For the hyperparameters, we set ${\lambda _v=1}$, ${\lambda _n=1}$, ${\lambda _e=1}$, ${\lambda _l=50}$ in our implementation.

\textbf{3D Pose Loss.} ${{{\mathcal{L}}_{\mathcal{J}}} = \sum\nolimits_{j = 1}^J {\left\| {{\bm{\phi}} _j^{3D} - \hat {\bm{\phi}} _j^{3D}} \right\|_2^2} }$, where ${{\bm{\phi}} _j^{3D}}$ and ${\hat {\bm{\phi}} _j^{3D}}$ are the ground truth and estimated 3D joint locations, respectively.

In our implementation, we first train the stacked hourglass network and the 3D pose regressor separately with the heat-map loss and the 3D pose loss, respectively. Then, we train the stacked hourglass network, the residual network and the Graph CNN for mesh generation with the combined loss ${{\mathcal{L}}_{fully}}$:
\begin{equation}
{{\mathcal{L}}_{fully}} = {\lambda _{\mathcal{H}}}{{\mathcal{L}}_{\mathcal{H}}} + {\lambda _{\mathcal{M}}}{{\mathcal{L}}_{\mathcal{M}}} + {\lambda _{\mathcal{J}}}{{\mathcal{L}}_{\mathcal{J}}},
\label{eq:fully}
\end{equation}
where ${\lambda _{\mathcal{H}}=0.5}$, ${\lambda _{\mathcal{M}}=1}$, ${\lambda _{\mathcal{J}}=1}$.

\subsection{Weakly-supervised Fine-tuning}
\label{Methodology:Training_real}
On the real-world dataset, \textit{i.e.}, the Stereo Hand Pose Tracking Benchmark~\cite{zhang20163d}, there is no ground truth of 3D hand mesh. Thus, we fine-tune the networks in a weakly-supervised manner. Moreover, our model also supports the fine-tuning without the ground truth of 3D joint \mbox{locations}, which can further removes the burden of annotating 3D joint locations on training data and make it more applicable for large-scale real-world dataset.

\textbf{Depth Map Loss.} As shown in Fig.~\ref{fig:Fig_Framework} (b), we leverage the reference depth map, which can be easily captured by a depth camera, as a weak supervision, and employ a differentiable renderer, similar to~\cite{kato2018renderer}, to render the estimated 3D hand mesh to a depth map from the camera viewpoint. We use smooth L1 loss~\cite{girshick2015fast} for the depth map loss:
\begin{equation}
{{\mathcal{L}}_{\mathcal{D}}} = smoot{h_{L1}}\left( {D,\hat D} \right),~\hat {\mathcal{D}} = {\mathcal{R}}\left( {\hat {\mathcal{M}}} \right),
\end{equation}
where ${\mathcal{D}}$ and ${\hat {\mathcal{D}}}$ denote the ground truth and rendered depth maps, respectively; ${\mathcal{R}\left( \cdot \right)}$ is the depth rendering function; ${\hat {\mathcal{M}}}$ is the estimated 3D hand mesh. We set the resolution of a depth map as 32$\times$32 px.

\begin{figure}[t]
	\FigTabBelowcaptionskip
	\begin{center}
		\includegraphics[width=1.0\linewidth]{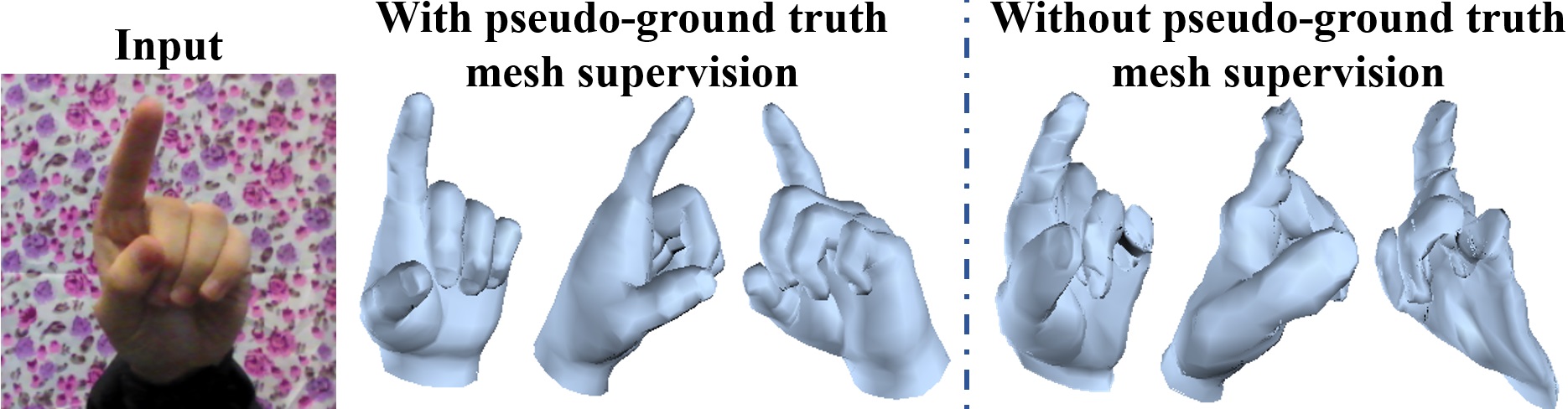}
		\vspace{-8pt}
		\caption{Impact of the pseudo-ground truth mesh supervision. Without the supervision of pseudo-ground truth mesh, the network produces very rough meshes with incorrect shape and noisy surface.}
		\label{fig:Fig_pseudo_mesh_supervision}
	\end{center}
\end{figure}

\textbf{Pseudo-Ground Truth Mesh Loss.} Training with only the depth map loss could lead to a degenerated solution, as shown in Fig.~\ref{fig:Fig_pseudo_mesh_supervision} (right), since the depth map loss only constrains the visible surface and is sensitive to the noise in the captured depth map. To solve this issue, inspired by~\cite{li2017learning}, we create the pseudo-ground truth mesh ${\tilde {\mathcal{M}}}$ by testing on the real-world training data using the pretrained models and the ground truth heat-maps. The pseudo-ground truth mesh ${\tilde {\mathcal{M}}}$ usually has reasonable edge length and good surface \mbox{smoothness}, although it suffers from the relative depth error. Based on this observation, we do not apply vertex loss or normal loss, and we only {adopt} the edge loss ${{\mathcal{L}}_{e}}$ and the Laplacian loss ${{\mathcal{L}}_{l}}$ as the pseudo-ground truth mesh loss ${{{\mathcal{L}}_{p{\mathcal{M}}}} = {\lambda _e}{{\mathcal{L}}_{e}} + {\lambda _l}{{\mathcal{L}}_{l}}}$, where ${\lambda _e=1}$, ${\lambda _l=50}$, in order to preserve the edge length and surface smoothness of the mesh. As shown in Fig.~\ref{fig:Fig_pseudo_mesh_supervision} (middle), with the supervision of the pseudo-ground truth meshes, the network can generate meshes with correct shape and smooth surface.

In our implementation, we first fine-tune the stacked hourglass network with the heat-map loss, and then end-to-end fine-tune all networks with the combined loss ${{\mathcal{L}}_{weakly}}$:
\begin{equation}
{{\mathcal{L}}_{weakly}} = {\lambda _{\mathcal{H}}}{{\mathcal{L}}_{\mathcal{H}}} + {\lambda _{\mathcal{D}}}{{\mathcal{L}}_{\mathcal{D}}} + {\lambda _{p{\mathcal{M}}}}{{\mathcal{L}}_{p{\mathcal{M}}}},
\label{eq:weakly}
\end{equation}
where ${\lambda _{\mathcal{H}}=0.1}$, ${\lambda _{\mathcal{D}}=0.1}$, ${\lambda _{p{\mathcal{M}}}=1}$. Note that Eq.~\ref{eq:weakly} is the loss function for fine-tuning on the dataset without 3D pose supervision. When the ground truth of 3D joint locations is provided during training, we add the 3D pose loss ${{\mathcal{L}}_{\mathcal{J}}}$ in the loss function and set the weight ${\lambda _{\mathcal{J}}=10}$.

%

%

\section{Experiments}
\label{Experiments}

\subsection{Datasets, Metrics and Implementation Details}
\label{Experiments:Datasets_metrics}
In this work, we evaluate our method on two aspects: 3D hand mesh reconstruction and 3D hand pose estimation.

For 3D hand mesh reconstruction, we evaluate the generated 3D hand meshes on our proposed synthetic and real-world {datasets}, which are introduced in Section~\ref{Methodology:dataset}, since no other hand RGB image dataset contains the ground truth of 3D hand meshes. We measure the average error in Euclidean \mbox{space} between the corresponding vertices in each generated 3D mesh and its ground truth 3D mesh. This metric is denoted as ``mesh error'' in the following experiments.

For 3D hand pose estimation, we evaluate our proposed methods on two publicly available datasets: Stereo Hand Pose Tracking Benchmark (STB)~\cite{zhang20163d} and the Rendered Hand Pose Dataset (RHD)~\cite{zimmermann2017learning}. STB is a real-world dataset containing 18,000 images with the ground truth of 21 3D hand joint locations and corresponding depth images. Following~\cite{zimmermann2017learning,cai2018weakly,spurr2018cross}, we split the dataset into 15,000 training samples and 3,000 test samples. To make the joint definition consistent with our settings and RHD dataset, following~\cite{cai2018weakly}, we move the root joint location from palm center to wrist. RHD is a synthetic dataset containing 41,258 training images and 2,728 testing images. This dataset is challenging due to the large variations in viewpoints and the low image resolution. We evaluate the performance of 3D hand pose estimation with three metrics: (i) Pose error: the average error in Euclidean space between the estimated 3D joints and the ground truth joints; (ii) 3D PCK: the percentage of correct keypoints of which the Euclidean error distance is below a threshold; (iii) AUC: the area under the curve on PCK for different error thresholds.

We implement our method within the PyTorch framework. The networks are trained using the RMSprop optimizer~\cite{tieleman2012lecture} with mini-batches of size 32. The learning rate is set as ${10^{-3}}$ when pretraining on our synthetic dataset, and is set as ${10^{-4}}$ when fine-tuning on RHD~\cite{zimmermann2017learning} and STB~\cite{zhang20163d}. The input image is resized to 256${\times}$256 px. Following the same condition used in~\cite{zimmermann2017learning,cai2018weakly,spurr2018cross}, we assume that the global hand scale and the absolute depth of root joint are provided at test time. The global hand scale is set as the length of the bone between MCP and PIP joints of the middle finger.

\subsection{Ablation Study of Loss Terms}
\label{Experiments:Ablation_study}

We first evaluate the impact of different losses used in the fully-supervised training (Eq.~\ref{eq:fully}) on the performance of mesh reconstruction and pose estimation. We conduct this experiment on our synthetic dataset. As presented in Table~\ref{tab:loss_compare}, the model trained with the full loss achieves the best performance in both mesh reconstruction and pose estimation, which indicates that all the losses have contributions to producing accurate 3D hand mesh as well as 3D hand joint locations.

\begin{table}[!t]
	\FigTabBelowcaptionskip
	\begin{center}\small
		\resizebox{1.0\linewidth}{!}{
		\begin{tabular}{cccccc}\hline
			Error (mm)&${-}$Normal&${-}$Edge&${-}$Laplacian&${-}$3D Pose&Full\\\hline
			Mesh error&8.34&9.09&8.63&9.04&\textbf{7.95}\\
			Pose error&8.30&9.06&8.55&9.24&\textbf{8.03}\\
			\hline
		\end{tabular}%
		}
		\vspace{-5pt}
		\caption{Ablation study by eliminating different loss terms from our fully-supervised training loss in Eq.~\ref{eq:fully}, respectively. We report the average mesh and pose errors evaluated on the validation set of our synthetic dataset.}
		\label{tab:loss_compare}%
	\end{center}
\end{table}%


\begin{figure}[t]
	\FigTabBelowcaptionskip
	\begin{center}
		\includegraphics[width=0.78\linewidth]{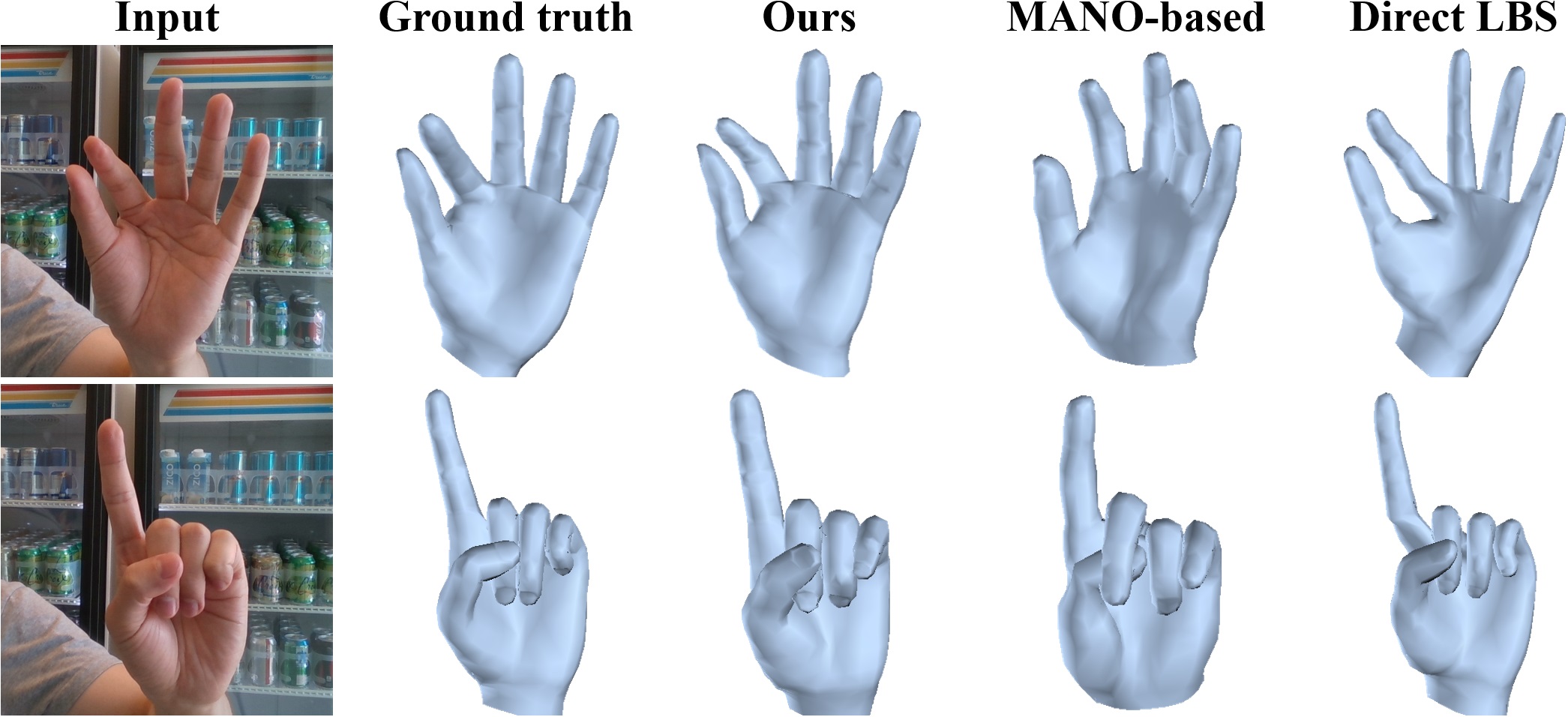}
		\vspace{-5pt}
		\caption{Qualitative comparisons of the meshes generated by our method and other methods. The meshes generated by the MANO-based method usually exhibit inaccurate shape and pose. The meshes generated by the direct Linear Blend Skinning (LBS) method suffer from serious artifacts. Examples are taken from our real-world dataset.}
		\label{fig:exp_compare_mesh}
	\end{center}
\end{figure}

\begin{table}[!t]
	\FigTabBelowcaptionskip
	\begin{center}\small
		\begin{tabular}{cccc}\hline
			Mesh error (mm) & MANO-based & Direct LBS & Ours\\\hline
			Our synthetic dataset & 12.12 & 10.32 & \textbf{8.01} \\
			Our real-world dataset & 20.86 & 13.33 & \textbf{12.72} \\
			\hline
		\end{tabular}%
		\vspace{-5pt}
		\caption{Average mesh errors tested on the validation set of our synthetic dataset and our real-world dataset. We compare our method with two baseline methods. Note that the mesh errors in this table are measured on the aligned mesh defined by MANO~\cite{romero2017embodied} for fair comparison.}
		\label{tab:mesh_compare}%
	\end{center}
\end{table}%


\subsection{Evaluation of 3D Hand Mesh Reconstruction}
\label{Experiments:Self_mesh}
We demonstrate the advantages of our proposed Graph CNN-based 3D hand mesh reconstruction method by comparing it with two baseline methods: direct Linear Blend Skinning (LBS) method and MANO-based method.


\textbf{Direct LBS.} We train the network to directly regress 3D hand joint locations from the heat-maps and the image features, which is similar to the network architecture proposed in~\cite{cai2018weakly}. We generate the 3D hand mesh from only the estimated 3D hand joint locations by applying inverse kinematics and LBS with the predefined mesh model and skinning weights (see the supplementary for \mbox{details}). As shown in Table~\ref{tab:mesh_compare}, the average mesh error of direct \mbox{LBS} method is worse than our method on both our synthetic dataset and our real-world dataset, since the \mbox{LBS} model for mesh generation is predefined and cannot be adapt to hands with \mbox{different} shapes. As can be seen in Fig.~\ref{fig:exp_compare_mesh}, the hand meshes generated by direct \mbox{LBS} method have unrealistic deformation at joints and suffer from serious inherent artifacts.

\textbf{MANO-based Method.} We also implement a MANO \cite{romero2017embodied} based method that regresses hand shape and pose parameters from the latent image features using three fully-connected layers. Then, the 3D hand mesh is generated from the estimated shape and pose parameters using MANO hand model~\cite{romero2017embodied} (see the supplementary for {details}). The networks are trained in fully-supervised manner using the same loss functions as Eq.~\ref{eq:fully} on our synthetic dataset. For fair comparison, we align our hand mesh with the MANO hand mesh, and compute mesh error on the aligned mesh. As shown in Table~\ref{tab:mesh_compare} and Fig.~\ref{fig:exp_compare_mesh}, the MANO-based method exhibits inferior performance on mesh reconstruction compared with our method. Note that direct supervising MANO parameters on synthetic dataset may obtain better performance~\cite{boukhayma20193d}. But it is infeasible on our synthetic dataset since our dataset does not contain MANO parameters.

\begin{figure}[t]
	\FigTabBelowcaptionskip
	\begin{center}
		\includegraphics[width=1.0\linewidth]{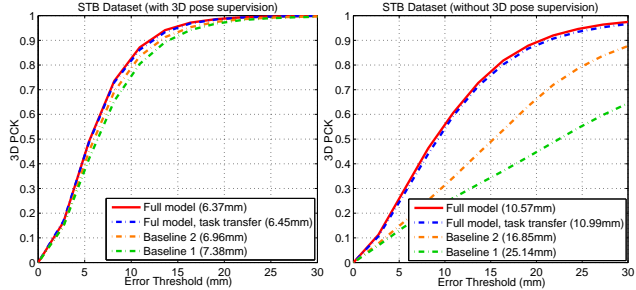}
		\vspace{-18pt}
		\caption{Self-comparisons of 3D hand pose estimation on STB dataset~\cite{zhang20163d}. \textbf{Left}: 3D PCK of the model fine-tuned with 3D hand pose supervision. \textbf{Right}: 3D PCK of the model fine-tuned without 3D hand pose supervision. The average pose errors are shown in parentheses.}
		\label{fig:exp_compare_self}
	\end{center}
\end{figure}

\begin{table}[!t]
	\FigTabBelowcaptionskip
	\begin{center}\small
		\begin{tabular}{ccc}\hline
			Method & Pipeline & Depth map loss \\\hline
			Baseline 1 & im${\rightarrow}$hm+feat${\rightarrow}$pose & \xmark \\
			Baseline 2 & im${\rightarrow}$hm+feat${\rightarrow}$mesh${\rightarrow}$pose & \xmark \\
			Full model & im${\rightarrow}$hm+feat${\rightarrow}$mesh${\rightarrow}$pose & \cmark\\
			\hline
		\end{tabular}%
		\FigTabBeforecaptionskip
		\caption{Differences between the baseline methods for 3D hand pose estimation and our full model.}
		\label{tab:self_compare}%
	\end{center}
\end{table}%

\subsection{Evaluation of 3D Hand Pose Estimation}
\label{Experiments:eval_pose}
We also evaluate our approach on the task of 3D hand pose estimation.

\begin{figure*}[t]
	\FigTabBelowcaptionskip
	\begin{center}
		\includegraphics[width=0.98\linewidth]{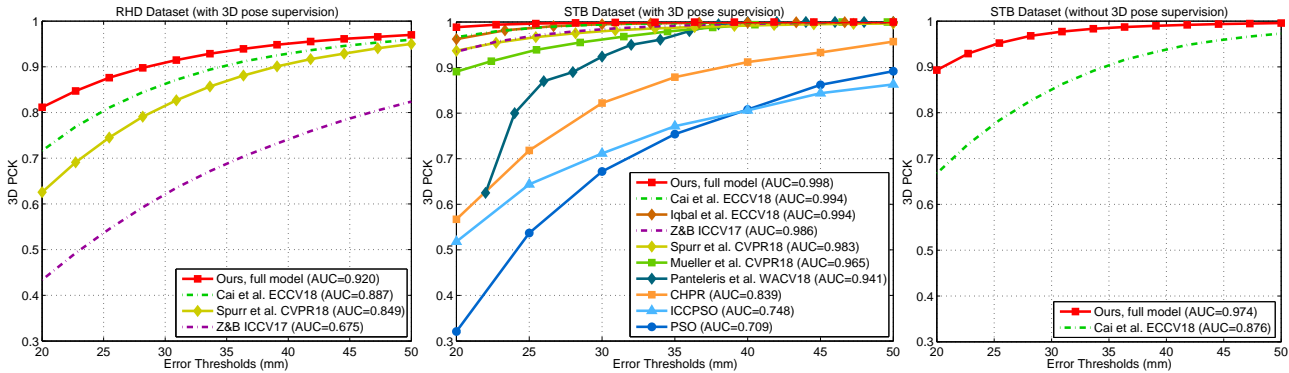}
		\vspace{-6pt}
		\caption{Comparisons with state-of-the-art methods on RHD~\cite{zimmermann2017learning} and STB~\cite{zhang20163d} dataset. \textbf{Left}: 3D PCK on RHD dataset~\cite{zimmermann2017learning} with 3D hand pose supervision. \textbf{Middle}: 3D PCK on STB dataset~\cite{zhang20163d} with 3D hand pose supervision. \textbf{Right}: 3D PCK on STB dataset~\cite{zhang20163d} without 3D hand pose supervision. The AUC values are shown in parentheses.}
		\label{fig:exp_compare_other}
	\end{center}
\end{figure*}

\textbf{Self-comparisons.} We conduct self-comparisons on STB dataset~\cite{zhang20163d} by fine-tuning the networks pretrained on our synthetic dataset in a weakly-supervised manner, as described in Section~\ref{Methodology:Training_real}. In Table~\ref{tab:self_compare}, we compare our proposed weakly-supervised method (\textbf{Full model}) with \mbox{two} baselines: (i) \textbf{Baseline 1}: directly regressing 3D hand joint locations from the heat-maps and the feature maps without using the depth map loss during training; (ii) \textbf{Baseline 2}: regressing 3D hand joint locations from the estimated 3D hand mesh without using the depth map loss during training. As presented in Fig.~\ref{fig:exp_compare_self}, the estimation accuracy of Baseline 2 is superior to that of Baseline 1, which indicates that our proposed 3D hand mesh reconstruction network is beneficial to 3D hand pose estimation. Furthermore, the estimation accuracy of our full model is superior to that of Baseline 2, especially when fine-tuning without 3D hand pose supervision, which validates the effectiveness of introducing the depth map loss as a weak supervision.

In addition, to explore a more efficient way for 3D hand pose estimation without mesh generation, we directly regress the 3D hand joint locations from the latent feature extracted by our full model instead of regressing them from the 3D hand mesh (see the supplementary for details). This task transfer method is denoted as ``\textbf{Full model, task transfer}'' in Fig.~\ref{fig:exp_compare_self}. Although this method has the same pipeline as that of Baseline 1, the estimation accuracy of this task transfer method is better than that of Baseline 1 and is only a little bit worse than that of our full model, which indicates that the latent feature extracted by our full model is more discriminative and is easier to regress accurate 3D hand pose than the latent feature extracted by Baseline 1.






\textbf{Comparisons with State-of-the-arts.} We compare our method with state-of-the-art 3D hand pose estimation methods on RHD~\cite{zimmermann2017learning} and STB~\cite{zhang20163d} datasets. The PCK curves over different error \mbox{thresholds} are presented in Fig.~\ref{fig:exp_compare_other}. On RHD dataset, as shown in Fig.~\ref{fig:exp_compare_other} (left), our method outperforms the three state-of-the-art methods~\cite{zimmermann2017learning,spurr2018cross,cai2018weakly} over all the error thresholds on this dataset. On STB dataset, when the 3D hand pose ground truth is given during training, we compare our methods with seven state-of-the-art \mbox{methods}~\cite{zhang20163d,zimmermann2017learning,panteleris2018using,spurr2018cross,mueller2018ganerated,cai2018weakly,iqbal2018hand}, and our method outperforms these methods over most of the error thresholds, as shown in Fig.~\ref{fig:exp_compare_other} (middle). We also experiment with the situation when 3D hand pose ground truth is unknown during training on STB dataset, and compare our method with the weakly-supervised method proposed by Cai \textit{et al}.~\cite{cai2018weakly}, both of which adopt reference depth maps as a weak supervision. As shown in Fig.~\ref{fig:exp_compare_other} (right), our 3D mesh-based method outperforms Cai \textit{et al}.~\cite{cai2018weakly} by a large margin.

\begin{figure}[t]
	\setlength{\belowcaptionskip}{-0.6cm}
	\begin{center}
		\includegraphics[width=1.0\linewidth]{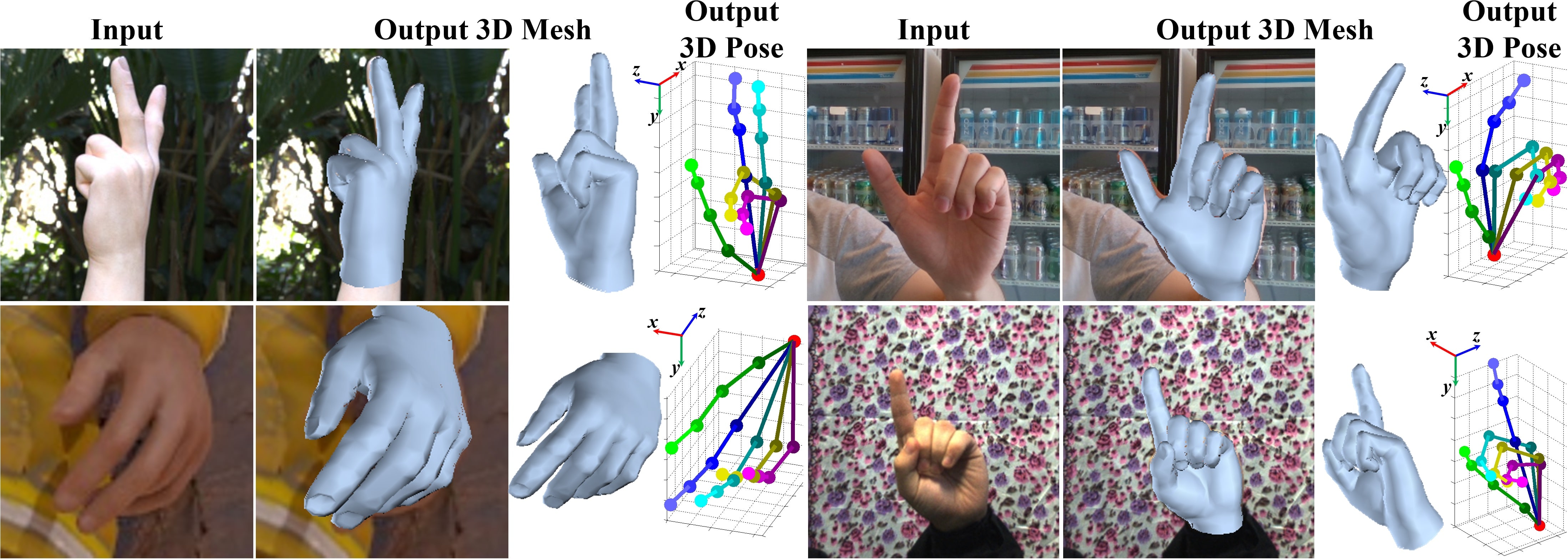}
		\vspace{-8pt}
		\caption{Qualitative results for our synthetic dataset (top left), our real-world dataset (top right), RHD dataset~\cite{zimmermann2017learning} (bottom left), and STB dataset~\cite{zhang20163d} (bottom right).}
		\label{fig:exp_qualitative}
	\end{center}
\end{figure}

\subsection{Runtime and Qualitative Results}
\label{Experiments:runtime_size}
\textbf{Runtime.} We evaluate the runtime of our method on one Nvidia GTX 1080 GPU. The runtime of our full model outputting both 3D hand mesh and 3D hand pose is 19.9ms on average, including 12.6ms for the stacked hourglass network forward propagation, 4.7ms for the residual network and Graph \mbox{CNN} forward propagation, and 2.6ms for the forward propagation of the pose regressor. Thus, our method can run in real-time on GPU at over 50fps.

\textbf{Qualitative Results.} Some qualitative results of 3D hand mesh reconstruction and 3D hand pose estimation for our synthetic dataset, our real-world dataset, RHD~\cite{zimmermann2017learning}, and STB~\cite{zhang20163d} datasets are shown in Fig.~\ref{fig:exp_qualitative}. More qualitative results are presented in the supplementary.

\section{Conclusion}
In this paper we have tackled the challenging task of 3D hand shape and pose estimation from a single RGB image. We have developed a Graph CNN-based model to reconstruct a full 3D mesh of hand surface from an input RGB image. To train the model, we have created a large-scale synthetic RGB image dataset with ground truth \mbox{annotations} of both 3D joint locations and 3D hand meshes, on which we train our model in a fully-supervised manner. To fine-tune our model on real-world datasets without 3D ground truth, we render the generated 3D mesh to a depth map and leverage the observed depth map as a weak supervision. Experiments on our proposed new datasets and two public datasets show that our method can recover accurate 3D hand mesh and 3D joint locations in real-time.

In future work, we will use Mocap data to create a larger 3D hand pose and shape dataset. We will also consider the cases of hand-object and hand-hand interactions in order to make the hand pose and shape estimation more robust.

\vspace{8pt}
\textbf{Acknowledgment:} This work is in part supported by MoE Tier-2 Grant (2016-T2-2-065) of Singapore. This work is also supported in part by start-up grants from \mbox{University} at Buffalo and a gift grant from Snap Inc.

{\small
	\bibliographystyle{ieee_fullname}
	\bibliography{egbib}

\begin{thebibliography}{10}\itemsep=-1pt

\bibitem{Arnold}
Autodesk.
\newblock Arnold renderer.
\newblock \url{https://www.arnoldrenderer.com}, 2018.

\bibitem{Maya}
Autodesk.
\newblock Maya.
\newblock \url{https://www.autodesk.com.sg/products/maya}, 2018.

\bibitem{bogo2016keep}
Federica Bogo, Angjoo Kanazawa, Christoph Lassner, Peter Gehler, Javier Romero,
  and Michael~J Black.
\newblock Keep it smpl: Automatic estimation of 3d human pose and shape from a
  single image.
\newblock In {\em ECCV}, 2016.

\bibitem{boukhayma20193d}
Adnane Boukhayma, Rodrigo de Bem, and Philip~HS Torr.
\newblock 3d hand shape and pose from images in the wild.
\newblock {\em CVPR}, 2019.

\bibitem{cai2018weakly}
Yujun Cai, Liuhao Ge, Jianfei Cai, and Junsong Yuan.
\newblock Weakly-supervised 3d hand pose estimation from monocular rgb images.
\newblock In {\em ECCV}, 2018.

\bibitem{chung1997spectral}
Fan~RK Chung and Fan~Chung Graham.
\newblock {\em Spectral graph theory}, volume~92.
\newblock American Mathematical Society, 1997.

\bibitem{de2011model}
Martin de La~Gorce, David~J Fleet, and Nikos Paragios.
\newblock Model-based 3d hand pose estimation from monocular video.
\newblock {\em IEEE Transactions on Pattern Analysis and Machine Intelligence},
  33(9):1793--1805, 2011.

\bibitem{defferrard2016convolutional}
Micha{\"e}l Defferrard, Xavier Bresson, and Pierre Vandergheynst.
\newblock Convolutional neural networks on graphs with fast localized spectral
  filtering.
\newblock In {\em NIPS}, 2016.

\bibitem{dhillon2007weighted}
Inderjit~S Dhillon, Yuqiang Guan, and Brian Kulis.
\newblock Weighted graph cuts without eigenvectors a multilevel approach.
\newblock {\em IEEE Transactions on Pattern Analysis and Machine Intelligence},
  29(11), 2007.

\bibitem{Flickr}
Flickr.
\newblock Flickr.
\newblock \url{https://www.flickr.com/}, 2018.

\bibitem{ge2018hand}
Liuhao Ge, Yujun Cai, Junwu Weng, and Junsong Yuan.
\newblock Hand pointnet: 3d hand pose estimation using point sets.
\newblock In {\em CVPR}, 2018.

\bibitem{ge2016Robust}
Liuhao Ge, Hui Liang, Junsong Yuan, and Daniel Thalmann.
\newblock Robust 3{D} hand pose estimation in single depth images: from
  single-view \mbox{CNN} to multi-view {CNNs}.
\newblock In {\em CVPR}, 2016.

\bibitem{ge2017_3D}
Liuhao Ge, Hui Liang, Junsong Yuan, and Daniel Thalmann.
\newblock {3D} convolutional neural networks for efficient and robust hand pose
  estimation from single depth images.
\newblock In {\em CVPR}, 2017.

\bibitem{Ge2018Real}
Liuhao Ge, Hui Liang, Junsong Yuan, and Daniel Thalmann.
\newblock Real-time {3D} hand pose estimation with 3d convolutional neural
  networks.
\newblock {\em IEEE Transactions on Pattern Analysis and Machine Intelligence},
  2018.

\bibitem{Ge2018Robust}
Liuhao Ge, Hui Liang, Junsong Yuan, and Daniel Thalmann.
\newblock Robust 3d hand pose estimation from single depth images using
  multi-view cnns.
\newblock {\em IEEE Transactions on Image Processing}, 27(9):4422--4436, 2018.

\bibitem{ge2018point}
Liuhao Ge, Zhou Ren, and Junsong Yuan.
\newblock Point-to-point regression pointnet for 3d hand pose estimation.
\newblock In {\em ECCV}, 2018.

\bibitem{girshick2015fast}
Ross Girshick.
\newblock Fast r-cnn.
\newblock In {\em ICCV}, 2015.

\bibitem{he2016deep}
Kaiming He, Xiangyu Zhang, Shaoqing Ren, and Jian Sun.
\newblock Deep residual learning for image recognition.
\newblock In {\em CVPR}, 2016.

\bibitem{RealSense}
Intel.
\newblock Intel realsense.
\newblock \url{https://realsense.intel.com/}, 2018.

\bibitem{iqbal2018hand}
Umar Iqbal, Pavlo Molchanov, Thomas Breuel, Juergen Gall, and Jan Kautz.
\newblock Hand pose estimation via latent 2.5 d heatmap regression.
\newblock In {\em ECCV}, 2018.

\bibitem{joseph2016fits}
David Joseph~Tan, Thomas Cashman, Jonathan Taylor, Andrew Fitzgibbon, Daniel
  Tarlow, Sameh Khamis, Shahram Izadi, and Jamie Shotton.
\newblock Fits like a glove: Rapid and reliable hand shape personalization.
\newblock In {\em CVPR}, 2016.

\bibitem{kanazawa2018end}
Angjoo Kanazawa, Michael~J Black, David~W Jacobs, and Jitendra Malik.
\newblock End-to-end recovery of human shape and pose.
\newblock In {\em CVPR}, 2018.

\bibitem{kato2018renderer}
Hiroharu Kato, Yoshitaka Ushiku, and Tatsuya Harada.
\newblock Neural 3d mesh renderer.
\newblock In {\em CVPR}, 2018.

\bibitem{Khamis2015learning}
Sameh Khamis, Jonathan Taylor, Jamie Shotton, Cem Keskin, Shahram Izadi, and
  Andrew Fitzgibbon.
\newblock Learning an efficient model of hand shape variation from depth
  images.
\newblock In {\em CVPR}, 2015.

\bibitem{lassner2017unite}
Christoph Lassner, Javier Romero, Martin Kiefel, Federica Bogo, Michael~J
  Black, and Peter~V Gehler.
\newblock Unite the people: Closing the loop between 3d and 2d human
  representations.
\newblock In {\em CVPR}, 2017.

\bibitem{li2017learning}
Zhizhong Li and Derek Hoiem.
\newblock Learning without forgetting.
\newblock In {\em ECCV}, 2017.

\bibitem{Liang2019Hough}
H. {Liang}, J. {Yuan}, J. {Lee}, L. {Ge}, and D. {Thalmann}.
\newblock Hough forest with optimized leaves for global hand pose estimation
  with arbitrary postures.
\newblock {\em IEEE Transactions on Cybernetics}, 49(2):527--541, 2019.

\bibitem{lin2014microsoft}
Tsung-Yi Lin, Michael Maire, Serge Belongie, James Hays, Pietro Perona, Deva
  Ramanan, Piotr Doll{\'a}r, and C~Lawrence Zitnick.
\newblock Microsoft coco: Common objects in context.
\newblock In {\em ECCV}, 2014.

\bibitem{loper2015smpl}
Matthew Loper, Naureen Mahmood, Javier Romero, Gerard Pons-Moll, and Michael~J
  Black.
\newblock {SMPL}: A skinned multi-person linear model.
\newblock {\em ACM Transactions on Graphics (TOG)}, 34(6):248, 2015.

\bibitem{makris2015model}
Alexandros Makris and A Argyros.
\newblock Model-based 3d hand tracking with on-line hand shape adaptation.
\newblock {\em BMVC}, 2015.

\bibitem{malik2018deephps}
Jameel Malik, Ahmed Elhayek, Fabrizio Nunnari, Kiran Varanasi, Kiarash
  Tamaddon, Alexis Heloir, and Didier Stricker.
\newblock Deephps: End-to-end estimation of 3d hand pose and shape by learning
  from synthetic depth.
\newblock In {\em 3DV}, 2018.

\bibitem{mueller2018ganerated}
Franziska Mueller, Florian Bernard, Oleksandr Sotnychenko, Dushyant Mehta,
  Srinath Sridhar, Dan Casas, and Christian Theobalt.
\newblock {GANerated} hands for real-time 3d hand tracking from monocular
  {RGB}.
\newblock In {\em CVPR}, 2018.

\bibitem{mueller2017real}
Franziska Mueller, Dushyant Mehta, Oleksandr Sotnychenko, Srinath Sridhar, Dan
  Casas, and Christian Theobalt.
\newblock Real-time hand tracking under occlusion from an egocentric {RGB-D}
  sensor.
\newblock In {\em ICCV}, 2017.

\bibitem{newell2016stacked}
Alejandro Newell, Kaiyu Yang, and Jia Deng.
\newblock Stacked hourglass networks for human pose estimation.
\newblock In {\em ECCV}, 2016.

\bibitem{oikonomidis2011efficient}
Iason Oikonomidis, Nikolaos Kyriazis, and Antonis Argyros.
\newblock \mbox{Efficient} model-based 3{D} tracking of hand articulations
  using \mbox{Kinect}.
\newblock In {\em BMVC}, 2011.

\bibitem{panteleris2018using}
Paschalis Panteleris, Iason Oikonomidis, and Antonis Argyros.
\newblock Using a single rgb frame for real time 3d hand pose estimation in the
  wild.
\newblock In {\em WACV}, 2018.

\bibitem{pavlakos2018learning}
Georgios Pavlakos, Luyang Zhu, Xiaowei Zhou, and Kostas Daniilidis.
\newblock Learning to estimate 3d human pose and shape from a single color
  image.
\newblock {\em CVPR}, 2018.

\bibitem{rad2018domain}
Mahdi Rad, Markus Oberweger, and Vincent Lepetit.
\newblock Domain transfer for 3d pose estimation from color images without
  manual annotations.
\newblock In {\em ACCV}, 2018.

\bibitem{ranjan2018generating}
Anurag Ranjan, Timo Bolkart, Soubhik Sanyal, and Michael~J Black.
\newblock Generating 3d faces using convolutional mesh autoencoders.
\newblock {\em ECCV}, 2018.

\bibitem{rehg1994visual}
James~M Rehg and Takeo Kanade.
\newblock Visual tracking of high dof articulated structures: an application to
  human hand tracking.
\newblock In {\em ECCV}, 1994.

\bibitem{remelli2017low}
Edoardo Remelli, Anastasia Tkach, Andrea Tagliasacchi, and Mark Pauly.
\newblock Low-dimensionality calibration through local anisotropic scaling for
  robust hand model personalization.
\newblock In {\em ICCV}, 2017.

\bibitem{romero2017embodied}
Javier Romero, Dimitrios Tzionas, and Michael~J Black.
\newblock Embodied hands: Modeling and capturing hands and bodies together.
\newblock {\em ACM Transactions on Graphics (TOG)}, 36(6):245, 2017.

\bibitem{simon2017hand}
Tomas Simon, Hanbyul Joo, Iain~A Matthews, and Yaser Sheikh.
\newblock Hand keypoint detection in single images using multiview
  bootstrapping.
\newblock In {\em CVPR}, 2017.

\bibitem{spurr2018cross}
Adrian Spurr, Jie Song, Seonwook Park, and Otmar Hilliges.
\newblock Cross-modal deep variational hand pose estimation.
\newblock In {\em CVPR}, 2018.

\bibitem{sridhar2016real}
Srinath Sridhar, Franziska Mueller, Michael Zollh{\"o}fer, Dan Casas, Antti
  Oulasvirta, and Christian Theobalt.
\newblock Real-time joint tracking of a hand manipulating an object from rgb-d
  input.
\newblock In {\em ECCV}, 2016.

\bibitem{sridhar2013interactive}
Srinath Sridhar, Antti Oulasvirta, and Christian Theobalt.
\newblock Interactive markerless articulated hand motion tracking using rgb and
  depth data.
\newblock In {\em ICCV}, 2013.

\bibitem{stenger2006model}
Bj{\"o}rn Stenger, Arasanathan Thayananthan, Philip~HS Torr, and Roberto
  Cipolla.
\newblock Model-based hand tracking using a hierarchical bayesian filter.
\newblock {\em IEEE Transactions on Pattern Analysis and Machine Intelligence},
  28(9):1372--1384, 2006.

\bibitem{tan2017indirect}
J Tan, Ignas Budvytis, and Roberto Cipolla.
\newblock Indirect deep structured learning for 3d human body shape and pose
  prediction.
\newblock In {\em BMVC}, 2017.

\bibitem{taylor2014user}
Jonathan Taylor, Richard Stebbing, Varun Ramakrishna, Cem Keskin, Jamie
  Shotton, Shahram Izadi, Aaron Hertzmann, and Andrew Fitzgibbon.
\newblock User-specific hand modeling from monocular depth sequences.
\newblock In {\em CVPR}, 2014.

\bibitem{tieleman2012lecture}
Tijmen Tieleman and Geoffrey Hinton.
\newblock Lecture 6.5-rmsprop, coursera: Neural networks for machine learning.
\newblock {\em University of Toronto, Technical Report}, 2012.

\bibitem{tkach2017online}
Anastasia Tkach, Andrea Tagliasacchi, Edoardo Remelli, Mark Pauly, and Andrew
  Fitzgibbon.
\newblock Online generative model personalization for hand tracking.
\newblock {\em ACM Transactions on Graphics (TOG)}, 36(6):243, 2017.

\bibitem{tompson2014real}
Jonathan Tompson, Murphy Stein, Yann Lecun, and Ken Perlin.
\newblock Real-time continuous pose recovery of human hands using convolutional
  networks.
\newblock {\em ACM Transactions on Graphics (ToG)}, 33(5):169, 2014.

\bibitem{tung2017self}
Hsiao-Yu Tung, Hsiao-Wei Tung, Ersin Yumer, and Katerina Fragkiadaki.
\newblock Self-supervised learning of motion capture.
\newblock In {\em NIPS}, 2017.

\bibitem{varol2018bodynet}
G{\"u}l Varol, Duygu Ceylan, Bryan Russell, Jimei Yang, Ersin Yumer, Ivan
  Laptev, and Cordelia Schmid.
\newblock Bodynet: Volumetric inference of 3d human body shapes.
\newblock {\em ECCV}, 2018.

\bibitem{verma2018feastnet}
Nitika Verma, Edmond Boyer, and Jakob Verbeek.
\newblock Feastnet: Feature-steered graph convolutions for 3d shape analysis.
\newblock In {\em CVPR}, 2018.

\bibitem{wang2018pixel2mesh}
Nanyang Wang, Yinda Zhang, Zhuwen Li, Yanwei Fu, Wei Liu, and Yu-Gang Jiang.
\newblock Pixel2mesh: Generating 3d mesh models from single rgb images.
\newblock In {\em ECCV}, 2018.

\bibitem{wu2001hand}
Ying Wu and Thomas~S Huang.
\newblock Hand modeling, analysis and recognition.
\newblock {\em IEEE Signal Processing Magazine}, 18(3):51--60, 2001.

\bibitem{wu2005analyzing}
Ying Wu, John Lin, and Thomas~S Huang.
\newblock Analyzing and capturing articulated hand motion in image sequences.
\newblock {\em IEEE transactions on pattern analysis and machine intelligence},
  27(12):1910--1922, 2005.

\bibitem{yan2018spatial}
Sijie Yan, Yuanjun Xiong, and Dahua Lin.
\newblock Spatial temporal graph convolutional networks for skeleton-based
  action recognition.
\newblock {\em arXiv preprint arXiv:1801.07455}, 2018.

\bibitem{yu15lsun}
Fisher Yu, Yinda Zhang, Shuran Song, Ari Seff, and Jianxiong Xiao.
\newblock Lsun: Construction of a large-scale image dataset using deep learning
  with humans in the loop.
\newblock {\em arXiv preprint arXiv:1506.03365}, 2015.

\bibitem{yuan2018depth}
Shanxin Yuan, Guillermo Garcia-Hernando, Bj{\"o}rn Stenger, Gyeongsik Moon, Ju
  Yong~Chang, Kyoung Mu~Lee, Pavlo Molchanov, Jan Kautz, Sina Honari, Liuhao
  Ge, et~al.
\newblock Depth-based 3d hand pose estimation: From current achievements to
  future goals.
\newblock In {\em CVPR}, 2018.

\bibitem{zhang20163d}
Jiawei Zhang, Jianbo Jiao, Mingliang Chen, Liangqiong Qu, Xiaobin Xu, and
  Qingxiong Yang.
\newblock {3D} hand pose tracking and estimation using stereo matching.
\newblock {\em arXiv preprint arXiv:1610.07214}, 2016.

\bibitem{zimmermann2017learning}
Christian Zimmermann and Thomas Brox.
\newblock Learning to estimate 3d hand pose from single {RGB} images.
\newblock In {\em ICCV}, 2017.

\end{thebibliography}
}

\newpage

\appendix
\section*{Supplementary}
\section{Qualitative Results}
We present more qualitative results of 3D hand mesh reconstruction and 3D hand pose estimation for our synthetic dataset, our real-world dataset, STB dataset~\cite{zhang20163d}, RHD dataset~\cite{zimmermann2017learning}, and Dexter+Object dataset~\cite{sridhar2016real}, as shown in Fig.~\ref{fig:qualitative_result}. Please see the supplementary video for more qualitative results on continuous sequences.

\begin{figure*}[!b]
	\FigTabBelowcaptionskip
	\begin{center}
		\includegraphics[width=1.0\textwidth]{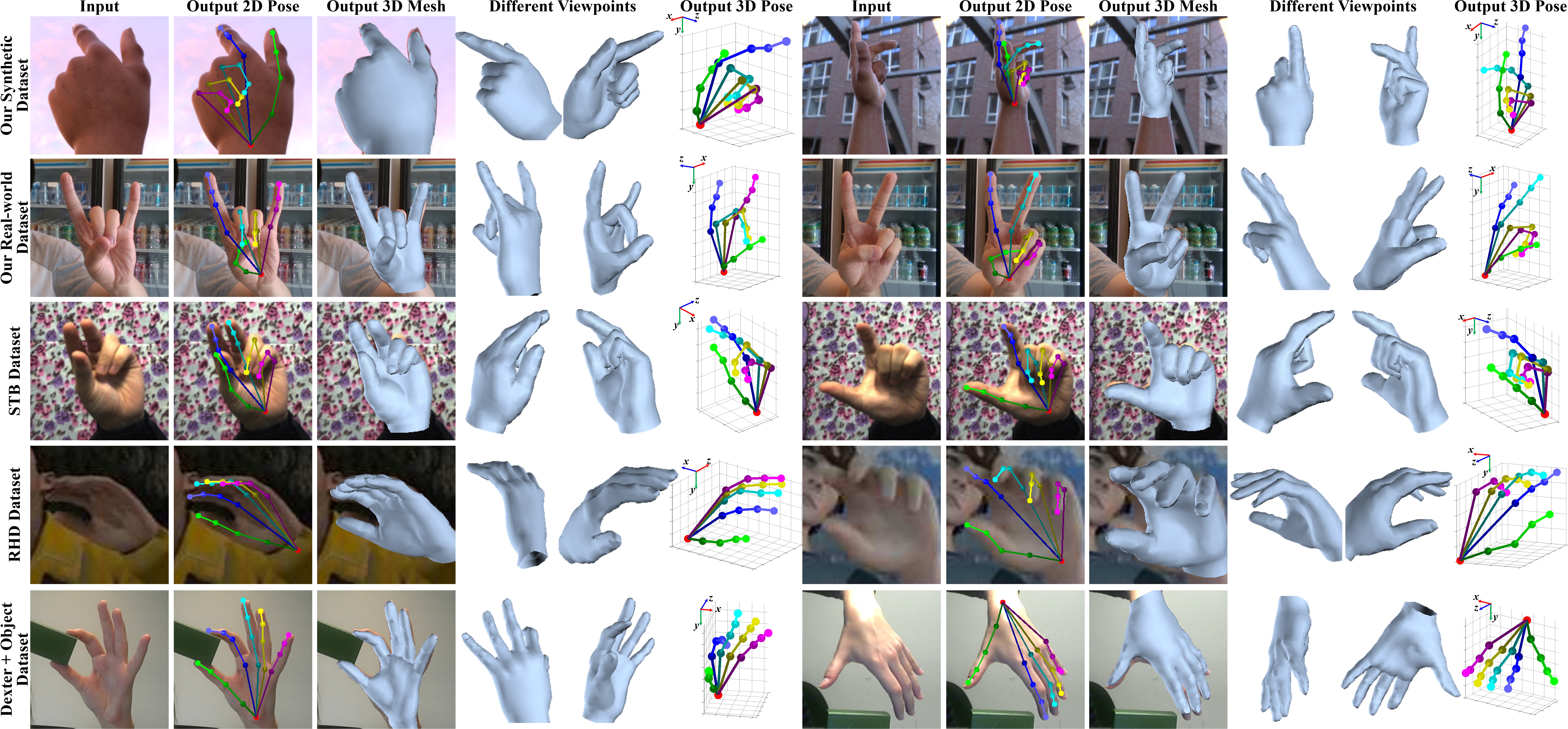}
		\FigTabBeforecaptionskip
		\caption{Qualitative results for our synthetic dataset (the first row), our real-world dataset (the second row), STB dataset~\cite{zhang20163d} (the third row), RHD dataset~\cite{zimmermann2017learning} (the fourth row), and Dexter+Object dataset~\cite{sridhar2016real} (the last row).}
		\label{fig:qualitative_result}
	\end{center}
\end{figure*}

\section{Details of Baseline Methods for 3D Hand Mesh Reconstruction}
\begin{figure*}[!t]
	\FigTabBelowcaptionskip
	\begin{center}
		\includegraphics[width=0.8\linewidth]{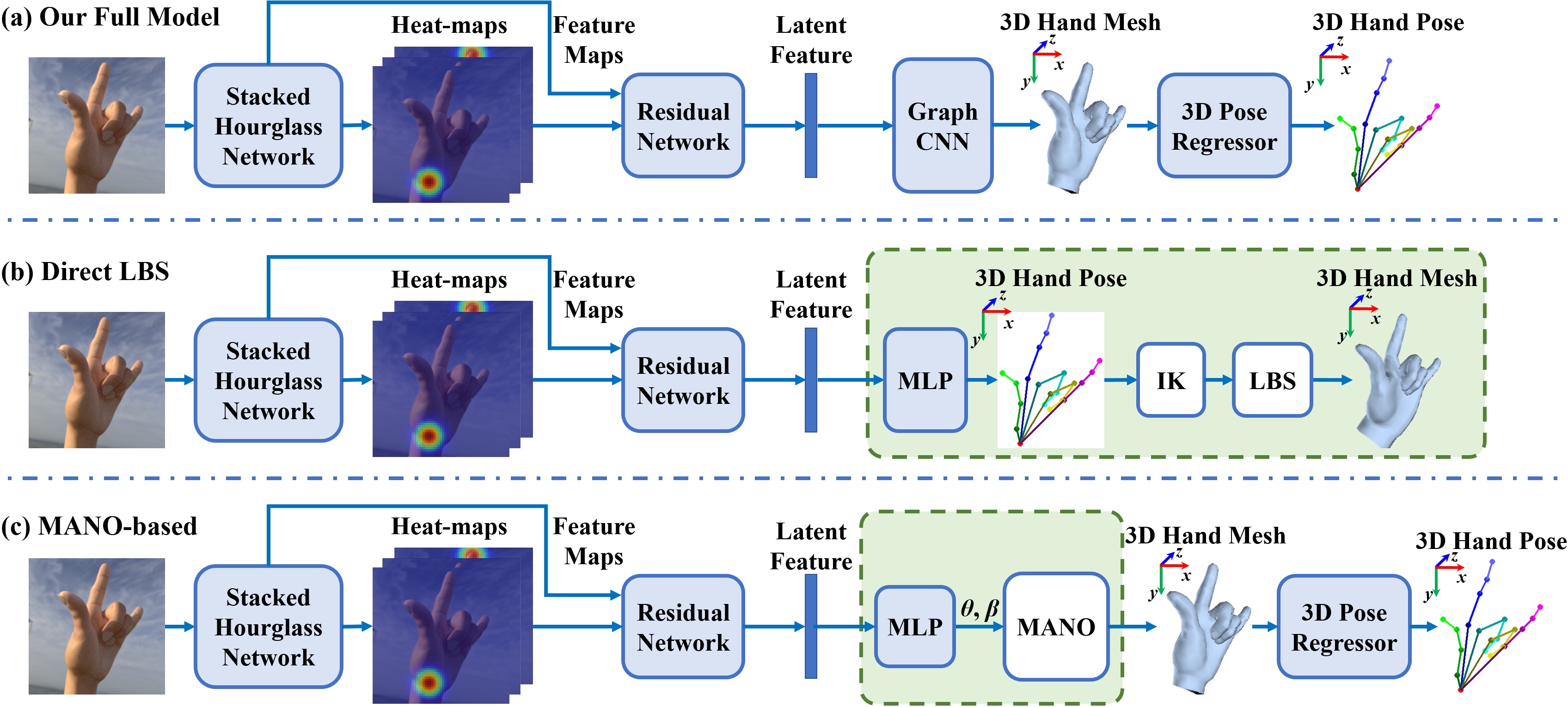}
		\FigTabBelowcaptionskip
		\caption{Pipelines of our proposed method and two baseline methods: direct LBS method and MANO-based method. The differences between the two baseline methods and our proposed method are highlighted in the green dashed line box.}
		\label{fig:lbs_mano}
	\end{center}
\end{figure*}

In Section 5.3 of our main paper, we compare our proposed method with two baseline methods for 3D hand mesh reconstruction: direct Linear Blend Skinning (LBS) method and MANO-based method. Here, we describe more details of these two baseline methods, as illustrated in Fig.~\ref{fig:lbs_mano}.

In the direct LBS method, we train the network to regress 3D hand joint locations from the heat-maps and the image features with heat-map loss and 3D pose loss. As illustrated in Fig.~\ref{fig:lbs_mano} (b), the latent feature extracted from the input image is mapped to 3D hand joint locations through a multi-layer perceptron (MLP) network with three fully-connected layers. Then, we apply inverse kinematics (IK) to compute the transformation matrix of each hand joint from the the estimated 3D hand joint locations. The 3D hand mesh is generated by applying LBS with the predefined hand model and skinning {weights}. In this method, the 3D hand mesh is only determined by the estimated 3D hand joint locations, thus it cannot be adapted to various hand shapes. In addition, the IK often suffers from singularity and multiple solutions, which makes the solutions to transformation matrices unreliable. Experimental results in Figure~7 and Table~2 of our main paper have shown the limitations of this direct LBS method.

In the MANO-based method, we train the network to regress hand shape and pose parameters of the MANO hand model~\cite{romero2017embodied}. As illustrated in Fig.~\ref{fig:lbs_mano} (c), the latent feature extracted from the input image is mapped to hand shape and pose parameters ${\theta}$, ${\beta}$ through an MLP network with three fully-connected layers. Then, the 3D hand mesh is generated from the regressed parameters ${\theta}$, ${\beta}$ using the MANO hand model~\cite{romero2017embodied}. Note that the MANO mesh generation module is differentiable and is involved in the network training. The networks are trained with heat-map loss, mesh loss and 3D pose loss, which are the same as our method. Since the MANO hand model is fixed during training and is essentially LBS with blend shapes~\cite{romero2017embodied}, the representation power of this method is limited. Experimental results in Figure~7 and Table~2 of our main paper have shown the limitations of this MANO-based method.

\section{Details of the Task Transfer Method}
\begin{figure*}[!t]
	\FigTabBelowcaptionskip
	\begin{center}
		\includegraphics[width=0.75\linewidth]{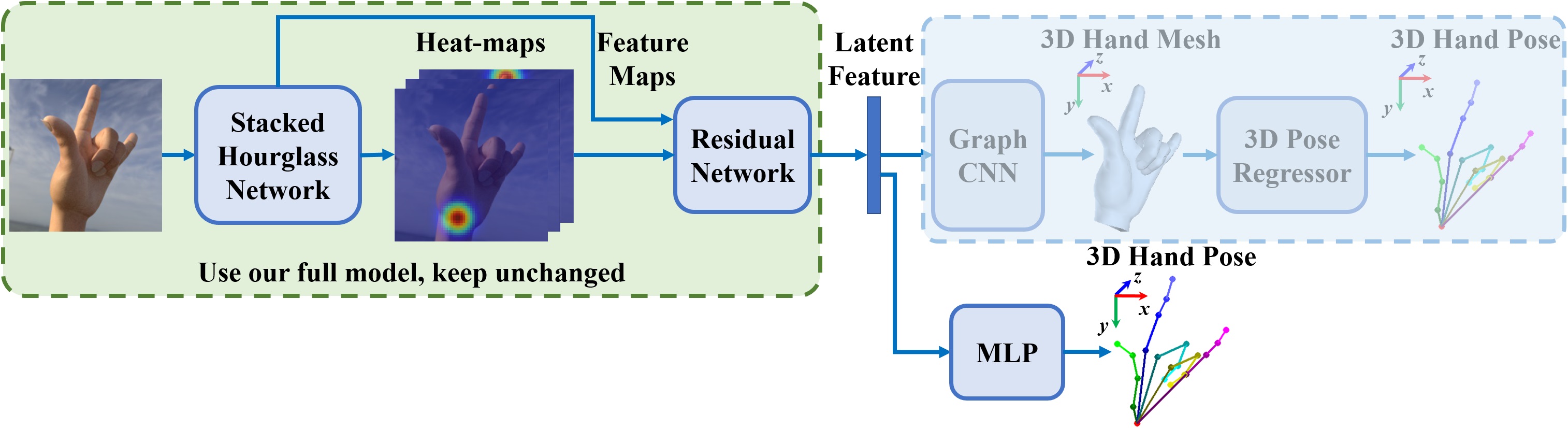}
		\FigTabBelowcaptionskip
		\caption{Illustration of our ``full model, task transfer'' method. We transfer our full model trained for 3D hand mesh reconstruction to the task of 3D hand pose estimation. Note that when training for the task of 3D hand pose estimation, the stacked hourglass network and the residual network are keep unchanged with our full model which is fully trained for the task of 3D hand mesh reconstruction.}
		\label{fig:task_transfer}
	\end{center}
\end{figure*}

In Section 5.4 of our main paper, we implement an alternative method (``full model, task transfer'') for 3D hand pose estimation by transferring our full model trained for 3D hand mesh reconstruction to the task of 3D hand pose estimation. Here, we describe more details of our task transfer method. As illustrated in Fig.~\ref{fig:task_transfer}, we directly regress the 3D hand joint locations from the latent feature extracted by our full model using an MLP network with three fully-connected \mbox{layers}. We first train the MLP network with 3D pose loss on our synthetic dataset. When experimenting on STB dataset~\cite{zhang20163d} with 3D pose supervision, we fine-tune the MLP network with 3D pose loss. When experimenting on STB dataset~\cite{zhang20163d} without 3D pose supervision, we directly use the MLP network pretrained on our synthetic dataset. Experimental results in Figure~8 of our main paper show that our task transfer method is better than the baseline method which is only trained for 3D hand pose estimation, even though these two methods have the same pipeline. This indicates that the latent feature extracted by our full model is more discriminative and is easier to regress accurate 3D hand pose since our full model is trained with the dense supervision of the 3D hand mesh that contains richer information than the 3D hand pose. In addition, although the estimation accuracy of our task transfer method is a little bit worse than that of our full model, our task transfer method is faster than our full model, since it does not generate 3D hand mesh. The runtime of our task transfer method is 15.1ms, while the runtime of our full model which estimate 3D hand pose from hand mesh is 19.9ms. Thus, in applications that only require 3D hand pose estimation but not 3D hand shape estimation, we can choose to use this task transfer method, which can maintain a comparable accuracy as our full model while runs at faster speed.

\end{document}